\documentclass[12pt]{article}
\usepackage[utf8]{inputenc}

\newcommand{\blind}{1}

\usepackage{amsfonts,amsmath,amsthm,amssymb,color,graphicx,bm,bbm}
\usepackage{enumerate}
\usepackage{physics}
\usepackage{enumitem}
\usepackage{caption}
\usepackage{multirow}
\usepackage{enumitem}
\usepackage{natbib}
\usepackage{booktabs}
\usepackage{mathtools}
\usepackage{setspace}
\usepackage{subcaption}
\usepackage[flushleft]{threeparttable} 
\usepackage{booktabs,caption}
\usepackage[ruled,linesnumbered]{algorithm2e}

\usepackage{soul}
\definecolor{darkpowderblue}{rgb}{0.0, 0.2, 0.6}
\definecolor{goldenpoppy}{rgb}{0.99, 0.76, 0.0}
\definecolor{cardinal}{rgb}{0.77, 0.12, 0.23}
\usepackage{hyperref}
\hypersetup{
    colorlinks=true,
    linkcolor=cardinal,
    citecolor=darkpowderblue,      
    urlcolor=cyan,
}

\usepackage{cleveref}

\newcommand{\mc}[1]{\mathcal{#1}}

\newcommand{\mb}[1]{\mbox{\textbf{#1}}}
\newcommand{\reals}{{\mathbb{R}}}
\makeatletter
\newcommand{\pushright}[1]{\ifmeasuring@#1\else\omit\hfill$\displaystyle#1$\fi\ignorespaces}
\newcommand{\pushleft}[1]{\ifmeasuring@#1\else\omit$\displaystyle#1$\hfill\fi\ignorespaces}
\makeatother

\newtheorem{theorem}{Theorem}
\newtheorem*{theorem*}{Theorem}
\newtheorem{thm}{Theorem}

\newtheorem{remark}{Remark}

\newtheorem{proposition}{Proposition}
\newtheorem{corollary}{Corollary}[theorem]

\newcommand{\dVt}{\Delta_{\mV_t}}
\newcommand{\dUt}{\Delta_{\mU_t}}

\newcommand{\dalphat}{\Delta_{\valpha_t}}

\newcommand{\drhot}{\Delta_{\rho_t}}

\newcommand{\proj}{\mathcal{P}_{{\mPsi_{[q]}}}}

\usepackage{amsmath,amsfonts,bm}
\newcommand{\widebar}[1]{\overline{#1}}

\def\eqref#1{equation~\ref{#1}}

\def\1{\bm{1}}

\def\vzero{{\bm{0}}}
\def\vone{{\bm{1}}}

\def\vbeta{{\bm{\beta}}}

\def\valpha{{\bm{\alpha}}}
\def\vepsilon{{\bm{\epsilon}}}
\def\vvarphi{{\bm{\varphi}}}
\def\vphi{{\bm{\phi}}}

\def\vb{{\bm{b}}}
\def\vc{{\bm{c}}}

\def\ve{{\bm{e}}}

\def\vk{{\bm{k}}}

\def\vu{{\bm{u}}}
\def\vv{{\bm{v}}}
\def\vw{{\bm{w}}}
\def\vx{{\bm{x}}}

\def\mA{{\bm{A}}}

\def\mD{{\bm{D}}}

\def\mI{{\bm{I}}}
\def\mJ{{\bm{J}}}
\def\mK{{\bm{K}}}

\def\mM{{\bm{M}}}

\def\mO{{\bm{O}}}

\def\mU{{\bm{U}}}
\def\mV{{\bm{V}}}

\def\mY{{\bm{Y}}}

\def\mPhi{{\bm{\Phi}}}

\def\mTheta{{\bm{\Theta}}}
\def\mGamma{{\bm{\Gamma}}}
\def\mPsi{{\bm{\Psi}}}

\DeclareMathAlphabet{\mathsfit}{\encodingdefault}{\sfdefault}{m}{sl}
\SetMathAlphabet{\mathsfit}{bold}{\encodingdefault}{\sfdefault}{bx}{n}

\DeclareMathOperator*{\argmin}{arg\,min}

\allowdisplaybreaks

\begin{document}
\def\spacingset#1{\renewcommand{\baselinestretch}%
{#1}\small\normalsize} \spacingset{1}



\if1\blind
{
  \title{\bf  Knowledge-Embedded Latent Projection for Robust Representation Learning}
  \author{ Weijing Tang \\
    \normalsize Department of Statistics and Data Science, Carnegie Mellon University
	\\
    and \\
    Ming Yuan \\
   \normalsize Department of Statistics, Columbia University
   \\
   and \\
    Zongqi Xia \\
   \normalsize Department of Neurology, University of Pittsburgh
   \\
   and \\
    Tianxi Cai \\
   \normalsize Department of Biostatistics, Harvard University}
    \date{}
  \maketitle
} \fi

\if0\blind
{
  \bigskip
  \bigskip
  \bigskip
  \begin{center}
    {\LARGE\bf Knowledge-Embedded Latent Projection for Robust Representation Learning}
\end{center}
    \vspace{15mm}
  \medskip
} \fi

\bigskip
\begin{abstract}
Latent space models are widely used for analyzing high-dimensional discrete data matrices, such as patient-feature matrices in electronic health records (EHRs), by capturing complex dependence structures through low-dimensional embeddings. However, estimation becomes challenging in the imbalanced regime, where one matrix dimension is much larger than the other. In EHR applications, cohort sizes are often limited by disease prevalence or data availability, whereas the feature space remains extremely large due to the breadth of medical coding system. Motivated by the increasing availability of external semantic embeddings, such as pre-trained embeddings of clinical concepts in EHRs, we propose a knowledge-embedded latent projection model that leverages semantic side information to regularize representation learning. Specifically, we model column embeddings as smooth functions of semantic embeddings via a mapping in a reproducing kernel Hilbert space. We develop a computationally efficient two-step estimation procedure that combines semantically guided subspace construction via kernel principal component analysis with scalable projected gradient descent. We establish estimation error bounds that characterize the trade-off between statistical error and approximation error induced by the kernel projection. Furthermore, we provide local convergence guarantees for our non-convex optimization procedure. Extensive simulation studies and a real-world EHR application demonstrate the effectiveness of the proposed method.
\end{abstract}

\noindent%

\spacingset{1.8} 

\setlength{\abovedisplayskip}{2.5pt}
\setlength{\belowdisplayskip}{2.5pt}
\newpage
\section{Introduction}

High-dimensional discrete data arise in many domains, including variant mutation data in genomics, electronic health records (EHRs) in clinical data, and bipartite relational data in social science~\citep{ma2023generalized, li2020inferring, wu2024general}.
These data can be represented as asymmetric matrices, where the two dimensions correspond to different types of entities. 
Examples include patient-clinical feature matrices in EHR data, patient-variant mutation matrices in genomic studies, and paper-author matrices in co-authorship data.  Despite their high dimensionality, such data often exhibit structured dependence patterns across rows and columns that can be explained by a relatively small number of underlying latent factors.
A natural and widely used approach in this setting is latent space modeling via representation learning
\citep{van2017neural,kopf2021latent,lavravc2021representation}. These methods assign low-dimensional embeddings to row and column entities and model observed discrete outcomes through interaction between the corresponding latent representations.  
This modeling perspective provides a flexible yet interpretable framework for capturing complex dependency structures in high-dimensional discrete data, while simultaneously yielding low-dimensional embeddings useful for downstream tasks such as visualization, clustering of similar entities, risk profiling, and missing data imputation. \citep{hoff2002latent, chen2018robust,  ma2020universal, wu2024general}

Despite their success in many applications, existing latent space modeling approaches often fall short in a challenging yet common real-world data regime characterized by imbalanced matrix dimensions, where one dimension is much larger than the other. 
To ground our discussion, we use EHR data as a running example throughout this work, where the number of clinical features can far exceed the number of patients.
\vspace{-4mm}
\paragraph{The EHR Example.} 
Commonly used EHR data naturally take the form of high-dimensional binary observations, where each patient record consists of a large collection of clinical features that are either present or absent. These features include diagnosis and procedure codes, medication prescriptions, laboratory test indicators, and clinical concepts extracted from free-text notes~\citep{choi2018mime, yu2015toward}.
Such data can be represented as a binary patient–feature matrix, where rows correspond to patients and columns correspond to clinical features, and an entry indicates whether a particular feature occurs in a patient’s medical history. 
Due to the extensive granularity of modern medical coding systems, many EHR features  encode overlapping or closely related clinical semantics. 
Representing clinical features using low-dimensional vector embeddings has been shown to be effective for capturing this redundancy~\citep{choi2018mime, hong2021clinical, liu2024representation, gan2025arch}.
However, in many EHR applications, particularly rare-disease and disease-specific cohort studies, there exists an imbalance between the number of patients and the number of clinical features.
Cohort sizes are often limited by disease prevalence, inclusion criteria, or data availability, whereas the feature space remains extremely large due to the breadth of possible medical codes and extracted concepts. As a result, the patient–feature matrix is both highly imbalanced.

In this imbalanced regime, accurate estimation of latent space embeddings for both rows and columns becomes challenging.
For the standard generalized latent factor model (GLFM), the average estimation error scales as $\mc{O}((n+p)/np)$ up to logarithmic factors, where $n$ is the number of rows and $p$ is the number of columns~\citep{chen2019joint, wang2022maximum,chen2023statistical}.
This suggests that, when $n \ll p$ in the imbalanced regime, the estimation error is dominated by the limited sample size $n$.
This challenge is further compounded by the pervasive sparsity of observations in the high-dimensional binary matrix data~\citep{wu2024general}. In EHR data, only a small subset of all possible clinical features appear in each patient's record. This leads to a binary matrix that is both highly imbalanced and sparse, with most entries equal to zero.

At the same time, a key characteristic of the motivating example is that the column entities correspond to semantic objects, rather than abstract indices. 
In EHR data, clinical features such as diagnosis and procedure codes can be linked to pretrained embeddings learned from large biomedical corpora or large-scale observational data, which encode clinically meaningful relationships among features~\citep{hong2021clinical, gan2025arch}.
This availability of semantic side information presents an opportunity to regularize latent structure estimation: semantically similar features or tasks should have similar latent embeddings. Leveraging this structure can help mitigate the estimation challenge posed by imbalance and sparsity.

Motivated by this observation, we propose a knowledge-embedded latent projection (KELP) model for high-dimensional binary matrices that incorporates semantic side information into latent representation learning. 
Our model is closely related to GLFM, in which the probability of observing a one in each entry is determined by interactions between latent embeddings of row and column entities. 
Different from GLFM, we do not treat column embeddings as unrestricted free parameters. Instead, we assume they vary smoothly as functions of their pre-trained semantic embeddings, through a mapping within a Reproducing Kernel Hilbert Space (RKHS). 
This smoothness assumption induces a semantically guided low-dimensional column space, which reduces the effective number of parameters while preserving the interpretability of latent space models.
Our framework accommodates general non-linear mappings from external semantic embeddings to latent representations, with linear transformations as a special case~\citep{mikolov2013exploiting, Shi2021-xe}. 

We develop a computationally efficient estimation procedure based on a two-step strategy. 
First, we construct a semantically guided column subspace using kernel principal component analysis, which captures the dominant variation implied by the RKHS mapping. 
Second, we estimate the latent space model using a scalable projected gradient descent algorithm that constrains the column embeddings to lie within this learned  subspace throughout optimization. 
To accommodate practical scenarios where external semantic information may be weakly informative or mismatched with the observed data, we further introduce a data-driven kernel selection procedure that prevents negative knowledge fusion by adaptively selecting between candidate kernels and a baseline model that relies solely on the binary observations.

Our theoretical analysis of the proposed KELP estimator characterizes a trade-off between estimation error arising from randomness in the noisy observations and approximation error induced by potential mismatch between the semantic embeddings and the true latent column embeddings.
Our analysis demonstrates that incorporating semantic side information substantially improves the scaling of the estimation error with the cost of the approximation error. This offers advantages over the standard GLFM in imbalanced regimes where the number of features far exceeds the sample size.
Moreover, we analyze the optimization properties of our non-convex optimization procedure. We establish local convergence guarantees for the projected gradient descent algorithm, showing that, under suitable initialization, the iterates converge linearly to a neighborhood that attains the same statistical estimation rate as the global optimizer, up to a multiplicative factor determined by the condition number of the underlying signal matrix.

The rest of the paper is organized as follows. 
\Cref{sec: model} introduces the proposed KELP model and discusses identifiability. 
\Cref{sec: estimation} presents the estimation procedure along with the data-driven kernel selection method.
\Cref{sec: theory} established the estimation error bounds, which are further validated through extensive simulation studies in~\Cref{sec: simulation}. 
We further demonstrate the practical utility of the KELP approach through an application to EHR data in \Cref{sec: real data}.

\paragraph{Notation} 
For a vector \(\vb\), \(\operatorname{diag}(\vb)\) denotes the diagonal matrix whose diagonal entries are the elements of \(\vb\). 
For a matrix $\mA \in \reals^{n\times p}$, we use $\|\mA\|_F$, $\|\mA\|_{2}$, $\|\mA\|_*$ to denote the Frobenius norm, spectral norm, and nuclear norm respectively. 
We use $\kappa_{\mA}$ to denote the condition number of $\mA$, and use $\mA_{*j}$ and $\mA_{i*}$ to denote the $j$-th column and the $i$-th row of $\mA$, respectively. 
Let $\vone_p$ and $\vzero_p$ denote the $p$-dimensional vectors of all ones and all zeros, respectively. 
We use $\mathcal{O}(r)$ to denote the set of all $r \times r$ orthonormal matrices. 
For two sequences $\{a_n\}$ and $\{b_n\}$, we denote $b_n\lesssim a_n$ if there exist a universal constant $C$ such that $|b_n| \le C |a_n|$. We denote $b_n\asymp a_n$ if $b_n\lesssim a_n$ and $a_n\lesssim b_n$. 
\section{Knowledge-embeded Latent Projection (KELP) Model}
\label{sec: model}

We consider a high-dimensional binary data matrix $\mY = [y_{ij}]_{i \in [n], j \in [p]} \in \{0,1\}^{n\times p}$ arising from interactions between two types of entities.
In the EHR motivating example, rows index patients and columns index clinical features, with $y_{ij}=1$ indicating the presence of feature $j$ in patient $i$’s record. 
In disease-specific cohorts, especially for rare diseases such as multiple sclerosis, the data matrix is often high-dimensional and imbalanced, with the sample size $n$ often substantially smaller than the number of features~$p$. 
Nonetheless, the column entities are semantic objects with pre-trained meaningful external semantic representations, which we denote by embedding vectors $\ve_j \in \mathbb{R}^d$. This serves as side information that can inform the latent structure. 
This motivates a KELP model that captures low-dimensional interaction structure while explicitly incorporating the semantic embeddings of column entities.

Specifically, we embed each row entity $i$ and column entity $j$ into a shared $r$-dimensional latent space, with embeddings $\vu_i \in \mathbb{R}^r$ and $\vv_j \in \mathbb{R}^r$, respectively. Conditional on external semantic embeddings $\ve_j \in \mathbb{R}^d$ for the column entities, we model the entries $y_{ij}$ as independent Bernoulli variables with probabilities
\begin{align} \label{eq: glfm-kf}
	\mathcal{P}(y_{ij}=1 \mid \{\ve_j\}_{j \in [p]}) =\sigma(\rho+ \alpha_i +  \vu_i^\top \vv_j) \quad \text{with}\quad   \vv_j = \vvarphi(\ve_j), \text{ for  } i \in[n], j \in [p],
\end{align}
where $\sigma(x)=e^x/(1+e^x)$ is the sigmoid function that maps a real-valued number to a probability within $[0,1]$ and $\vvarphi=(\varphi_1,\ldots,\varphi_r):\mathbb{R}^d\rightarrow\mathbb{R}^r$ is a mapping from the semantic embedding $\ve_j$ to $\vv_j$. 
In the context of EHR data, given other parameters fixed, a higher inner-product $\vu_i^\top \vv_j$ corresponds to a higher probability that feature $j$ appears in patient $i$’s record. 
Thus, in the shared latent space, the presence of column entity $j$ for row entity $i$ is more likely when their latent representations are closely aligned in direction. Moreover, conditional on a row embedding $\vu_i$, column entities with similar embeddings $\vv_j$ tend to co-occur with similar probabilities. Beyond the inner-product term, the parameter $\alpha_i \in \mathbb{R}$ captures row-specific heterogeneity, with larger values indicating a larger probability for patient $i$ to exhibit more EHR features in a record. The global intercept $\rho \in \mathbb{R}$ controls the overall sparsity level of the data matrix and is allowed to diverge as $n,p \to \infty$, which accommodates settings in which positive entries are rare, such as low prevalence of clinical features in EHR data.

Moreover, in (\ref{eq: glfm-kf}), we model $\vv_j$ as smooth functions of the external semantic embeddings $\ve_j$ through an unknown mapping $\vvarphi=(\varphi_1,\ldots,\varphi_r):\mathbb{R}^d\rightarrow\mathbb{R}^r$, rather than treating them as unconstrained parameters.
This design reflects the intuition that column entities with similar semantic representations should have similar latent embeddings. 
We assume that each component function $\varphi_k$, for $k \in [r]$, belongs to a reproducing kernel Hilbert space (RKHS) $\mc{H}$ associated with a positive semi-definite kernel $\mc{K}(\cdot, \cdot ; \eta)$, where $\eta$ denotes potential tuning parameters for the kernel function. Different choices of $\mc{K}$ induce different notions of smoothness. Common examples include the Gaussian kernel $\mc{K}(\ve_1, \ve_2; \eta)=\exp(-\|\ve_1-\ve_2\|_2^2/2\eta^2)$, which captures smooth nonlinear dependence, and the polynomial kernel $\mc{K}(\ve_1, \ve_2;\eta)=(\ve_1^\top \ve_2+\eta)^d$, which captures $d$-way interaction and includes the linear kernel as a special case when $d=1$. The latter corresponds to $\vvarphi(\ve)$ being linear in~$\ve$, a setting studied in prior work on semantic embedding translation \citep{mikolov2013exploiting, Shi2021-xe}. For brevity, we suppress the dependence on $\eta$ when it is clear from the context.

While the RKHS $\mc{H}$ may be infinite-dimensional, our model depends on $\vvarphi$ only through its evaluations at the finite set of observed semantic embeddings $\{\ve_1, \dots, \ve_p\}$. 
As explained below, this implies that there is no loss of generality to restrict our attention to mappings that are linear combinations of kernels evaluated at $\ve_j$'s, i.e., $\varphi_k(\cdot) = \sum_{j=1}^p a_{jk} \mc{K}(\cdot, \ve_j)$ with $a_{jk} \in \mathbb{R}$ for $1\le k \le r$, thanks to the representer lemma~\citep{kimeldorf1970correspondence}. Specifically, any $\varphi_k\in\mc H$ can be decomposed into a component lying in the span of the kernel sections $\{\mathcal{K}(\cdot,\ve_j)\}_{j=1}^p$ and an orthogonal remainder.
Since the orthogonal component of $\varphi_k$ evaluates to zero at the observed points, it does not affect the data-generating process. 
This further implies that every column of $\mV = [\varphi_k(\ve_j)]_{j\in[p],k\in[r]}\in \mathbb{R}^{p \times r}$ lies in the column space of the Gram matrix. 
To ensure model identifiability, we require that the mappings be centered. This amounts to parameterizing $\mV$ as $\mV =\mK_c \mA$ with the doubly centered Gram matrix $\mK_c = \mJ_p \mK_p \mJ_p.$ Here, $\mK_p=[\mc{K}(\ve_j, \ve_{j'})]_{1\le j,j'\le p}$ is the Gram matrix, $\mJ_p = \mI_p - \frac{1}{p}\vone_p \vone_p^\top$ is the centering matrix, and $\mA=[a_{jk}] \in \mathbb{R}^{p \times r}$ is the coefficient matrix. 
It is important to note that, while $\mV = \mK _c\mA$ takes the form of a  linear reparameterization, it serves as a structural constraint rather than a simple change of basis. 
The use of the RKHS framework allows us to enforce smoothness with respect to the semantic embeddings.
Specifically, because the leading eigenvectors of the kernel matrix correspond to components with high smoothness, an appropriate low-rank approximation of this structure effectively reduces the degrees of freedom in the parameter space and results in smooth mappings.

For notational convenience, let $\valpha = (\alpha_1, \cdots, \alpha_n)^\top \in \reals^n$, $\mU=(\vu_1^\top, \dots, \vu_n^\top)^\top \in \reals^{n\times r}$, $\mV=(\vv_1^\top, \dots, \vv_p^\top)^\top \in \reals^{p\times r}$, $\mPsi = [\psi_{\ell}(\ve_j)]_{1\le j \le p, 1\le \ell \le \mc{M}} \in \reals^{p\times \mc{M}}$, and $\mGamma=[\gamma_{\ell}^{(k)} ]_{1\le \ell \le \mc{M}, 1\le k \le r} \in \reals^{\mc{M} \times r}$. Denote the logit-transformed probability by $\Theta_{ij}=\text{logit}(\mathcal{P}(y_{ij}=1 \mid \{\ve_j\}_{j \in [p]}) )$ and $\mTheta = [\Theta_{ij}]_{i \in [n], j \in [p]} \in \reals^{n \times p}$. Then the model  can be represented in a matrix form, i.e.,
\[\mTheta = \rho \vone_n \vone_p^\top + \valpha \vone_p^\top + \mU \mV^\top \text{ with } \mV=\mK_c \mA.\]

\paragraph{Identifiability.} Note that multiple sets of parameters $\{\rho, \valpha, \mU, \mV\}$  give the same data distribution due to two facts that (1) $\rho+ \alpha_i =(\rho+c)+ (\alpha_i-c) $ holds for any constant~$c$; and (2) $\mU \mV^\top   = \mU \mM \cdot \mM^{-1} \mV^\top  $ holds for any invertible matrix $\mM \in \reals^{r \times r}$. 
To address the identifiability issue, we impose the following constraints:\begin{align}\valpha^\top \vone_n = 0, \quad \mV^\top \vone_p = \vzero_r, \quad \text{and} \quad \mU^\top \mU = \mV^\top \mV. \label{eq: iden}\end{align}
The constraint $\mV^\top \vone_p = \vzero_r$ forces the column embeddings to be centered so that the inner-product term is orthogonal to the row heterogeneity parameters, which is satisfied by the choice of centered mapping. By parameterizing $\mV = \mK_c \mA$ using the doubly centered Gram matrix $\mK_c$, we have $\mV^\top \vone_p = \mA^\top \mK_c^\top \vone_p = \vzero_r$.
The constraint $\mU^\top \mU = \mV^\top \mV$ fixes the relative scaling of the row and column embeddings.
The following proposition establishes that under these constraints, the parameters in~(\ref{eq: glfm-kf}) are identifiable up to an orthogonal rotation. The proof is given in~Supplementary Material~A.
\begin{proposition}[Identifiability] \label{prop: iden}
Let $\mTheta = \{\rho, \valpha, \mU, \mV\}$ and $\widebar{\mTheta} = \{\widebar{\rho}, \widebar{\valpha}, \widebar{\mU}, \widebar{\mV}\}$ be two sets of parameters that specify the same conditional distribution in the model (\ref{eq: glfm-kf}). 
Assume that both sets satisfy the constraints in (\ref{eq: iden}) and that $\mU, \widebar{\mU}, \mV, \widebar{\mV}$ have full rank. 
Then there exists an orthogonal matrix $\mO \in \mathcal{O}(r)$ such that $ \rho = \bar{\rho}, \  \valpha = \widebar{\valpha}, \  \mU = \widebar{\mU} \mO, \ \text{and} \  \mV = \widebar{\mV} \mO. $
\end{proposition}

\begin{remark}[Comparison with generalized linear factor models] 
The key distinction between our KELP model and GLFM~\citep{chen2019joint, wang2022maximum,chen2023statistical} lies in how the column embeddings are specified.
Without knowledge fusion, the column embeddings $\vv_j$ are treated as unconstrained parameters, and our model reduces to a standard GLFM, where $\mU$ represents latent factors and $\mV$ corresponds to column loadings.
In this unconstrained setting, parameters can still be estimated relying solely on the binary data matrix. 
However, in the high-dimensional yet imbalanced regime where the number of columns $p$ is much larger than the number of rows $n$,  recent theory on GLFM~\citep{chen2019joint, wang2022maximum} suggests that the estimation accuracy of both factors and loadings is fundamentally bottlenecked by $n$. 
In contrast, our KELP method incorporates external semantic embeddings via a shared smooth mapping $\vvarphi(\cdot)$, which effectively constrains $\mV$ to a lower-dimensional space induced by semantic representations. This structure enables information sharing across columns, and leads to improved estimation, particularly in imbalanced regimes. These advantages are formalized in~\Cref{sec: theory}.
\end{remark}

\begin{remark}[Connection to low-rank regression models] 
When the mapping $\varphi(\cdot)$ is linear and the data matrix is transposed, our KELP model (\ref{eq: glfm-kf}) can be viewed as a high-dimensional multivariate marginal logistic regression. For column entities $j = 1,\dots,p$ with covariates $\ve_j \in \mathbb{R}^d$, we observe $n$ binary responses $(y_{1j}, \dots, y_{nj})$. The coefficient matrix $\mGamma \mU^\top $ is low-rank of rank $r$, implying that the regression coefficients associated with the $n$ responses lie in a common lower-dimensional subspace. 
This formulation is closely related to reduced-rank regression (RRR), which has been extensively studied for multivariate linear models with continuous responses~\citep{chen2012sparse,bunea2011optimal,she2019cross} and, more recently, for logistic regression with binary responses~\citep{park2024low}. 
Despite this connection, the modeling objectives differ fundamentally from those in our setting. 
RRR aims to characterize the relationship between covariates and multiple responses through the coefficient matrix, often under additional sparsity assumptions and with a focus on coefficient estimation and inference. 
In contrast, our goal is to recover latent low-dimensional embeddings $(\mU, \mV)$ for both row and column entities, with column embeddings $\mV$ assumed to vary smoothly as functions of external semantic embeddings. 
We do not require correct specification of the linear kernel and the parameters of interest are the latent embeddings rather than regression coefficients.
As a result, our estimation and theoretical analysis focus on how semantic information reduces effective dimensionality and improves latent embedding estimation in imbalanced matrix settings.
\end{remark}

\section{Estimation Method}
\label{sec: estimation}

To estimate the KELP model parameters, we minimize the negative log-likelihood under the structural constraints.  
Recall from Section~\ref{sec: model} that the column embeddings lie in the span of the kernel evaluations, i.e.,  $ \mV=\mK_c \mA$. While this parameterization captures the smoothness structure, optimizing the full coefficient matrix $\mA\in \mathbb{R}^{p \times r}$ is computationally intensive, particularly in the high-dimensional imbalanced regime (see~\Cref{rmk: computational} for detailed discussion).
To obtain a scalable and numerically stable estimation procedure, we approximate the embedding space using kernel principal component analysis (KPCA). 
KPCA provides a data-adaptive truncation of the kernel spectrum, which retains only the leading eigencomponents that capture the dominant variation in the RKHS~\citep{scholkopf1997kernel}.  
This approximation preserves the smoothness structure implied by the kernel and restricts the column embeddings to a low-dimensional subspace, which leads to a computationally efficient likelihood-based estimation procedure.

\subsection{Likelihood-based Estimation via Kernel-PCA Approximation}
We first identify an informative subspace for the column embeddings by performing KPCA on the semantic embeddings. 
Specifically, we first perform the eigen-decomposition on the centered kernel matrix $\mK_c = \sum_{\ell=1}^p \mu_\ell  {\boldsymbol{\phi}}_\ell  {\boldsymbol{\phi}}_\ell^\top$, with eigenvalues $\mu_1 \ge \dots \ge \mu_p \ge 0$ and  eigenvectors $ {\boldsymbol{\phi}}_\ell \in \mathbb{R}^p$.
We select the smallest $q$ such that $\sum_{\ell=1}^{q} \mu_\ell / \sum_{\ell=1}^p \mu_\ell \geq 1-\delta$ for a small $\delta > 0$, and construct the basis $ {\mPsi}_{[q]}={ {\mPhi}_{[q]}}  {\mD_{[q]}}^{1/2}$ with 
${ {\mPhi}_{[q]}}=( {\vphi}_1^\top, \ldots,  {\vphi}_{q}^\top)^\top \in \reals^{p \times q}$ and $ {\mD_{[q]}}=\operatorname{diag}\left\{\mu_1, \ldots, \mu_{q}\right\}\in \reals^{q \times q}$.
We then approximate the column embeddings as $\mV =  {{\mPsi}_{[q]}} {\mGamma}$, where ${\mGamma} \in \mathbb{R}^{q \times r}$ are the coefficients for the $q$ leading kernel principal components. 

Given the  KPCA approximation, we estimate parameters by minimizing the negative log-likelihood over the constrained parameter space.
We define the feasible parameter~space:
\begin{align}
\mathcal{F}\left(n,p,q, M_1, M_2\right) =\Big\{\mTheta \mid     \mTheta & =\rho \vone_n \vone_p^\top + \valpha \vone_p^\top + \mU  \mV^\top, \mV =   {\mPsi_{[q]}} \mGamma, \valpha^\top \vone_n = 0, \nonumber\\
& \|\valpha\|_{\infty},\max _{i \in [n]}\left\|\mU_{i\centerdot}\right\|^2, \|\mGamma\|_F^2  \leqslant M,  -M_1 \leqslant \rho  \leqslant-M_2 \Big\},\label{eq:para space}
\end{align}
where $M$ is a constant, and $M_1 >  M_2 > 0$ are allowed to grow with $(n,p)$ to accommodate sparsity in high-dimensional settings (see~\Cref{rmk: sparsity} for detailed discussion about the impact of sparsity on the estimation error).
We omit the identifiability condition $\mV^\top \vone_p = \vzero_r$ from $\mathcal{F}\left(n,p,q, M_1, M_2\right)$, because it is automatically satisfied by  construction. Since $ {\mPsi}$ consists of eigenvectors of the centered kernel matrix, we have $  {\mPsi}_{[q]}^\top \vone_p = \vzero_q$ and thereby $\mV^\top \vone_p = \mGamma^\top  {\mPsi_{[q]}}^\top \vone_p = \vzero_r $.

Our estimator is the solution to the following constrained optimization problem:
\begin{align}
	& \min_{\mTheta \in \mathcal{F}\left(n,p,q, M_1, M_2\right)}   \mc{L}\left(\mTheta \right):=  - \sum_{i=1}^n \sum_{j=1}^p \left\{Y_{i j} \Theta_{i j}+\log \left(1-\sigma\left(\Theta_{i j}\right)\right)\right\}.\label{eq: global opt}
\end{align}

\subsection{Scalable Projected Gradient Descent Algorithm}

The optimization problem in~(\ref{eq: global opt}) is non-convex due to the low-rank constraint imposed on $\mTheta$ through the feasible set $\mathcal{F}\left(n,p,q, M_1, M_2\right)$. As a result, computing a global optimum is generally intractable. 
To address this challenge, we develop a scalable projected gradient descent algorithm that operates directly on the row and column embeddings $\mU$ and~$\mV$.
However, optimizing (\ref{eq: global opt}) under the factored parameterization $(\valpha, \rho, \mU, \mV)$ introduces non-uniqueness due to scale indeterminacy. As discussed in~\Cref{sec: model}, this issue can be mitigated by enforcing a balancing condition $\mU^\top \mU  = \mV^\top \mV$, which ensures that the singular values of $\mU \mV^\top$ are evenly distributed between the two factors $\mU$ and $\mV$. This approach has also been adopted in~\citet{park2018finding}.
Rather than imposing the constraint directly, we encourage balanced factorizations by optimizing a regularized objective:
\begin{align}
	\min_{\rho, \valpha, \mU, \mV}\mc{L}^{R}(\rho, \valpha, \mU, \mV) := &  \mc{L}\left( \rho \vone_n \vone_p^\top + \valpha \vone_p^\top + \mU \mV^{\top}  \right) + \frac{1}{4}g(\mU, \mV), \label{eq: nonconvex obj}  \\
	 \text{ subject to } & \quad \mV \in col( {\mPsi_{[q]}}) \quad \text{and} \quad \valpha^\top \vone_n = 0. \nonumber
\end{align}
where $g(\mU, \mV)=\|\mU^\top \mU - \mV ^\top \mV\|_F^2$ penalizes scale imbalance. For any optimal solution $\widehat{\mathbf{\Theta}}$ of~(\ref{eq: global opt}), there exists a balanced factorization $(\widehat{\mathbf{U}}, \widehat{\mathbf{V}})$ with $g(\widehat{\mathbf{U}}, \widehat{\mathbf{V}}) = 0$, implying that  the regularized problem (\ref{eq: nonconvex obj}) yields the same global estimator $\widehat{\mathbf{\Theta}}$ as~(\ref{eq: global opt}).

We summarize the PGD procedure in~\Cref{algo: PGD}.  The algorithm begins by constructing the basis $ {\mPsi}_{[q]}$ via  KPCA. It then iterates between gradient descent updates for $(\rho, \valpha, \mU, \mV)$ and projection steps that enforce the constraints $\mV \in \mathrm{col}( {\mPsi_{[q]}})$ and $\valpha^\top \vone_n = 0$ at each iteration. Step sizes are chosen to reflect the scale of each parameter block:  $\eta_{\rho} = \eta / (np)$, $\eta_{\alpha} = \eta / p$, and $\eta_{u} = \eta_v = \eta / \|[\mU_0^\top, \mV_0^\top ]\|_{2}^2$. These choices follow standard scaling rules used in low-rank matrix optimization.
We initialize the algorithm using universal singular value thresholding \citep{Chatterjee2015-hs}, followed by a projection to the column space of $ {\mPsi}_{[q]}$. Full details of the initialization procedure are provided in~Algorithm~S1.

\begin{algorithm}
\SetAlgoLined
\KwIn{Data matrix $\mY\in \{0,1\}^{n\times p}$, centered kernel matrix $\mK_c \in \mathbb{R}^{p \times p}$, projection dimension $q$, latent space dimension $r$, iterations $T$, step size parameter $\eta$.}
(1) \textit{KPCA}: Compute the top-$q$ eigenvectors of $\mK_c$ to form the basis $ {\mPhi}_{[q]} \in \mathbb{R}^{p \times q}$ and construct projection matrix $\mathcal{P}_{ {\mPsi}_{[q]}} =  {\mPhi}_{[q]} {\mPhi}_{[q]}^\top$\;
(2) \textit{Initialization}: Initialize $(\rho_0, \valpha_0, \mU_0, \mV_0, \mTheta_0)$ using~Algorithm~S1\;
(3) \For{$t=0,\dots, T-1$}{
  $\rho_{t+1}=\rho_t-\tau_\rho \langle \sigma(\mTheta_t)-\mY, \vone_n \vone_p^\top \rangle$\;
  $ {\valpha}_{t+1}=\valpha_t-\tau_\alpha (\sigma(\mTheta_t)-\mY)\vone_p $ \;
  ${\mU}_{t+1}
  =\mU_t-\tau_u\left((\sigma(\mTheta_t)-\mY)\mV_t + \mU_t(\mU_t^\top \mU_t - \mV_t ^\top \mV_t)\right)$\;
  $ {\mV}_{t+1}
  =\mV_t-\tau_v  \left((\sigma(\mTheta_t)-\mY)^\top \mU_t -  \mV_t  (\mU_t^\top \mU_t - \mV_t ^\top \mV_t)\right)$\;
  $\valpha_{t+1}=\mJ_n  {\valpha}_{t+1}$\;
  $\mV_{t+1}= \mathcal{P}_{ {\mPsi}_{[q]}}  {\mV}_{t+1}$\;
  $\mTheta_{t+1} =\rho_{t+1} \vone_n \vone_p^\top + \valpha_{t+1} \vone_p^\top +  \mU_{t+1} \mV_{t+1}^{\top}$ \;
 }
 \caption{Projected Gradient Descent with Kernal PCA}
 \label{algo: PGD}
 \KwOut{$(\widehat{\rho}, \widehat{\valpha}, \widehat{\mU}, \widehat{\mV})=(\rho_T, \valpha_T, \mU_T, \mV_T)$}
\end{algorithm}

\begin{remark}[Complexity Comparison with Dual Formulation] \label{rmk: computational}
A natural alternative is to estimate the mapping $\vvarphi(\cdot)$ directly in the RKHS using a dual formulation based on the Representer Theorem~\citep{kimeldorf1970correspondence}. 
In the dual formulation, the parameters are estimated by minimizing the regularized negative log-likelihood: 
\begin{align*}
	& \min_{\rho, \valpha, \mU, \vvarphi}   \mc{L}\left( \rho \vone_n \vone_p^\top + \valpha \vone_p^\top + \mU \mV^{\top}  \right) + \lambda \|\vvarphi\|_{\mathcal{H}}^2,   \\
	 \text{ subject to } & \quad  \valpha^\top \vone_n = 0, \ \mU^\top \mU  = \mV^\top \mV,  \ \text{and} \ \vv_j = \vvarphi(\ve_j), \text{ for  } j \in [p]. 
\end{align*}
By the Representer Theorem, the optimal column embeddings lie in the span of the centered kernel evaluations, leading to the parameterization $\mV = \mK_c \mA$ with a coefficient matrix $\mA \in \reals^{p \times r}$.
The objective can be reparameterized as $\min_{\rho, \valpha, \mU, \mA} \mc{L}\left( \rho \vone_n \vone_p^\top + \valpha \vone_p^\top + \mU \mA^{\top} \mK_c  \right) + \lambda \operatorname{tr}(\mA^\top \mK_c \mA)$. 
Gradient-based updates for $\mA$ take the form
$\mK_c(\nabla_{\mV}\mc{L} + \lambda \mA)$, which requires multiplication by the $p \times p$ kernel matrix $\mK_c$ at every iteration. This leads to a per-iteration cost of $\mathcal{O}(npr + p^2 r)$ for updating $\mA$, which is dominated by $p^2r$ in the imbalanced regime and prohibitive when $p$ is large. 
In contrast, our approach performs a one-time  KPCA step to identify a low-dimensional subspace of dimension $q \ll p$ and then restricts optimization to this subspace. This reduces the per-iteration complexity of updating $\mV$ to $\mathcal{O}(npr + pqr)$, which enables scalable estimation in the high-dimensional imbalanced regime ($n \ll p$).
\end{remark}

\subsection{Data-driven Kernel Selection}\label{sec: data-driven}
We propose a data-driven kernel selection approach based on the observed data. In our setting, the kernel defines how external semantic embeddings are mapped into the latent space of the data-driven representations; its suitability therefore depends on both the degree of alignment between the semantic structure and the empirical data structure, as well as the complexity required to capture this relationship. To adaptively select an appropriate kernel,  we randomly hold out a subset of entries from the binary matrix $\mY$, where each entry is  independently included with probability $\pi$. 
Let $\mM$ denote the corresponding masking matrix, with $M_{ij}=1$ indicating that the $(i,j)$-th entry is included in the hold-out set.

For each candidate kernel and its associated hyperparameters, we fit the model by minimizing the objective function in (\ref{eq: nonconvex obj}) using only the observed entries (i.e., $\{ (i,j):M_{ij}=0\}$) and then evaluate the fitted model on the hold-out entries. 
In addition to the candidate kernels introduced in Section~\ref{sec: model}, we include a baseline that does not incorporate external semantic knowledge and relies solely on the data. 
This baseline estimates parameters by optimizing (\ref{eq: nonconvex obj}) without imposing the constraint $\mV \in col( {\mPsi_{[q]}})$, 
and may be preferable when the number of rows $n$ is comparable to $p$ or when external semantic information is poorly aligned with the data structure.
The kernel (or baseline) that achieves the lowest hold-out loss is selected. In practice, we set $\pi=0.1$, which consistently yield stable and optimal kernel selection in our numerical experiments (see ~\Cref{sec: simulation}).

\begin{remark}[Generalization to New Unseen Column Entities]
Although we adopt a fixed design perspective and treat the semantic embeddings of the observed column entities as fixed, the kernel function $\mc{K}(\cdot, \cdot)$ is defined over the entire semantic space. As a result, latent embeddings for previously unseen column entities can be obtained via the Nystr$\ddot{o}$m extension \citep{williams2000using}, without refitting the model or requiring any observed interaction data. This capability is particularly valuable for accommodating newly introduced clinical variables.
For a new column entity with semantic embedding $\ve_{\text{new}}$, we compute its kernel similarity vector to the observed embeddings $\vk_{\text{new}} = p^{-1}(\mathcal{K}(\ve_{\text{new}}, \ve_1), \dots, \mathcal{K}(\ve_{\text{new}}, \ve_p))^\top \in \mathbb{R}^p$. Let $\bar{\vk} \in \mathbb{R}^p$ denote the column means of the scaled training kernel matrix $\mK_p$. 
After centering, the  KPCA feature representation of the new entity is given by 
$
 {\boldsymbol{\psi}}_{\text{new}} =  {\mD_{[q]}}^{-1/2}  {\mPhi_{[q]}}^\top (\vk_{\text{new}} - \bar{\vk}) \in \mathbb{R}^{q}.
$
The corresponding latent embedding is obtained by applying the learned linear mapping from kernel features to the column embedding space: $\widehat{\vv}_{\text{new}} = \widehat{\mGamma}^\top   {\boldsymbol{\psi}}_{\text{new}}= \widehat{\mV}^\top  {\mPhi_{[q]}}  {\mD_{[q]}}^{-1} {\mPhi_{[q]}}^\top (\vk_{\text{new}} - \bar{\vk}) $, where $\widehat{\mGamma}$ is the least-squares estimator obtained by regressing $\widehat{\mV}$ on $ {\mPsi}_{[q]}$.
This embedding enables prediction of interaction probabilities with existing row entities without additional model fitting. 
In contrast, standard GLFMs treat column embeddings as free parameters, and thus require additional observations and model refitting when new entities are introduced.
\end{remark}

\section{Theoretical Results}
\label{sec: theory}
In this section, we establish non-asymptotic error bounds for the KELP estimator, the solution to~(\ref{eq: global opt}). Our analysis explicitly characterizes the trade-off between estimation error, which is reduced by projecting the column space onto the top kernel principal components, and approximation error induced by this dimension reduction.  This decomposition highlights the benefits of knowledge fusion in high-dimensional, imbalanced settings where $n \ll p$. We further analyze the optimization behavior of the proposed nonconvex PGD algorithm, showing that it converges linearly to a statistical tolerance level and achieves the same error rate as the global optimizer up to a condition number factor.

\subsection{Statistical Error}
We first analyze the statistical error for the global optimum of the constrained optimization problem in~(\ref{eq: global opt}). 
Let the true parameters underlying the data-generating process in~(\ref{eq: glfm-kf}) be $\mTheta^*:= \rho^* \vone_n \vone_p^\top + \valpha^* \vone_p^\top + \mU^* {\mV^*}^{\top }$ with $\mV^*=\mK_c \mA^*$. 
Since $\mTheta^*$ may not lie in the constrained parameter space $\mathcal{F}\left(n,p,q, M_1, M_2\right)$, restricting the column space to the top $q$ kernel principal components can introduce approximation error.
To characterize this effect, we define the intermediate parameter $\mV_q^*$ as the  projection of the true embeddings $\mV^*$ onto $\mPhi_{[q]}$: $\mV_q^* = \mPhi_{[q]} \mPhi_{[q]}^\top \mV^* =\mPsi_{[q]}\mGamma^*$ with $\mGamma^* = \mD_{[q]}^{-1/2}\mPhi_{[q]}^\top \mV^*$.
Let $\widehat{\mTheta}$ denote the solution to the constrained optimization problem in~(\ref{eq: global opt}). 
The following theorem establishes the statistical error bound for $\widehat{\mTheta}$, the proof of which is provided in~Supplementary Material~B.

\begin{theorem} \label{thm: statistical error}
Assume the following conditions hold: (C1) the true parameters satisfy that $\|\valpha^*\|_{\infty}\leqslant M$ with ${\valpha^*}^\top \vone_n = 0$, $\max _{i \in [n]}\left\|\mU_{i\centerdot }^*\right\|^2\leqslant M$,  $-M_1 \leqslant\rho^* \leqslant-M_2$, and $\Theta^*_{ij} \lesssim -M_2$; (C2) the mapping $\vvarphi$ is bounded in the RKHS norm, i.e., $\| \vvarphi(\cdot)\|_{\mc{H}} \leqslant C_{\vvarphi}$ for some constant $C_{\vvarphi}$, and the kernel is bounded, i.e., $\sup_{\ve \in \mathbf{E}}\mc{K}(\ve, \ve)\leqslant C_K$ for some constant $C_K$. 
Then, there exist constants $c$ and $C$ such that uniformly over the parameter space $\mathcal{F}\left(n, p, q, M_1, M_2\right)$ with probability at least $1-(n+q)^{-c}$, we have
\begin{align*}
	\|\widehat{\mTheta}-\mTheta^* \|_F^2 \leqslant  & C \left(\mathcal{E}_{n,q,r}^2 + \mathcal{A}_{n,p,q}^2 \right),
\end{align*}
where the estimation error $\mathcal{E}_{n,q}^2$ and the approximation error $\mathcal{A}_{n,p,q}^2$ are given by
\begin{align*}
	\mathcal{E}_{n,q,r}^2  & = e^{2M_1} r (n+q) \log (n+q) \cdot  \max\left\{e^{-M_2},\frac{\log (n+q)}{n+q}\right\},\\
	 \mathcal{A}_{n,p,q}^2  &= e^{2M_1} \|\mU^{*}\|_{2}^2 \mu_{q+1}.
\end{align*}
\end{theorem}

\Cref{thm: statistical error} decomposes the statistical error into an estimation error $\mathcal{E}_{n,q,r}^2$, arising from the randomness of $\mY$, and an approximation error $\mathcal{A}_{n,p,q}^2$ induced by the KPCA approximation.
When the parameter space is correctly specified, i.e.,   $\mV^*$ lies in the span of the top $q$ kernel principle components, we have $\mu_{q+1}=0$ and the approximation error vanishes. 
In this case, the error is driven solely by $\mathcal{E}_{n,q,r}^2$, and the mean squared error scales as:
\[ \frac{1}{np}\|\widehat{\mTheta}-\mTheta^* \|_F^2 \asymp \frac{r(n+q)\log(n+q)}{np}\cdot \max\left\{e^{-M_2},\frac{\log (n+q)}{n+q}\right\}. \]
This result demonstrates the benefit of knowledge fusion. 
For a standard GLFM without external semantic information, the estimation error scales as $\mc{O}((n+p)/np)$ up to some logarithmic factors for fixed rank $r$ under dense observations~\citep{chen2019joint, wang2022maximum,chen2023statistical}. 
It requires both $n$ and $p$ to diverge for consistency.
In the high-dimensional yet imbalanced regime ($n \ll p$), the estimation error for GLFM is dominated by the small sample size $n$.
By constraining the column space to ${\mPsi_{[q]}}$, we effectively replace $p$ with $q$ in the numerator. 
When $q$ is fixed or grows slower than $p$, this leads to an improved error bound in the imbalanced regime.

On the other hand, this improvement comes at the cost of approximation error when $\mV^*$ does not fully lie in the truncated subspace.
The approximation term $\mathcal{A}_{n,p,q}^2$ depends on the spectral decay of the kernel operator. 
When the row embeddings are i.i.d.\ random variables with mean zero and bounded variance, we have $\|\mU^{*}\|_2^2 \asymp n$. 
Recall that $\mu_{q+1}$ is the $(q+1)$-th eigenvalue of the Gram matrix. 
By concentration results in \citet{Koltchinskii2000-pz, Braun2006-rc}, the scaled eigenvalue $\mu_{q+1}/p$ converges to the population eigenvalue of the kernel function, denoted by $\lambda_{q+1}$. 
Then, the total MSE scales as:
\[ 
\frac{1}{np}\|\widehat{\mTheta}-\mTheta^* \|_F^2 \asymp e^{2M_1}\frac{r(n+q)\log(n+q)}{np}\cdot \max\left\{e^{-M_2},\frac{\log (n+q)}{n+q}\right\} + e^{2M_1} \lambda_{q+1}. 
\]
The decay of $\lambda_{q+1}$ depends on the underlying kernel smoothness. 
For kernels with polynomial decay (e.g., Sobolev kernels), the eigenvalues scale as $\lambda_q \asymp q^{-2\nu}$, while for kernels with exponential decay (e.g., radial basis function (RBF) kernels), $\lambda_q \asymp \exp(-\nu q^2)$. 
Faster decay allows accurate approximation with smaller $q$.

Overall, \Cref{thm: statistical error} characterizes the trade-off between the estimation error $\mathcal{E}_{n,q,r}^2$ and the approximation error $\mathcal{A}_{n,p,q}^2$. 
When the rank $r$ is fixed and the binary matrix is dense, the error bound simplifies to $\mc{O}((n+q)/np + \lambda_{q+1})$ with logarithmic factors omitted.
For kernels with polynomial decay, the optimal choice is $q \asymp (np)^{\frac{1}{2\nu+1}}$, which leads to the error bound $\mathcal{O}\left((np)^{-\frac{2\nu}{2\nu+1}}+ p^{-1}\right)$.
For kernels with exponential decay, the optimal $q$ scales logarithmically, $q \asymp \sqrt{\log(np)}$. 
When $n \gtrsim \log(p)$, this leads to a rate of ${\mathcal{O}}\left(p^{-1}\right)$ up to logarithmic factors. 
This indicates an improvement over the standard GLFM estimation error rate of $\mc{O}(n^{-1})$ in the imbalanced regime ($n \ll p$).

\begin{remark}[Impact of Sparsity]\label{rmk: sparsity}
Recall that in the parameter space $\mc{F}(n,p,q,M_1,M_2)$,  $M$ is a universal
universal, while the bounds $M_1 > M_2 > 0$ are allowed to diverge as
$n,p \to \infty$. This implies that the probability of observing feature $j$ in subject $i$  satisfies
\[
e^{-M_1}
\;\lesssim\;
\mc{P}(y_{ij}=1 \mid \{\ve_j\}_{j\in[p]})
= \sigma(\Theta_{ij})
\;\lesssim\;
e^{-M_2}.
\]
Therefore, $e^{-M_2}$ characterizes the sparsity level, i.e., the proportion of ones
in the binary matrix.
The impact of sparsity is reflected in the estimation error through the factor
$
e^{2M_1}\max\!\left\{e^{-M_2},\frac{\log(n+q)}{n+q}\right\}.
$
In relatively dense regimes where $M_1 \asymp M_2$ and
$e^{-M_2} \gtrsim \frac{\log(n+q)}{n+q}$, the factor scales as
$e^{2M_1-M_2} \asymp e^{M_1}$.
This suggests that increased sparsity leads to larger statistical error.
Similar sparsity effects have been observed in latent space models for sparse binary symmetric matrix in statistical network analysis~\citep{ma2020universal}.

In the imbalanced regime $n \ll p$, sparsity leads to  larger estimation error for the standard GLFM than for our method, as demonstrated in the simulation results in~\Cref{subsec: simu sparsity}. 
By constraining the column space using external semantic information, our knowledge-fused method reduces the dimensionality from $p$ to $q$ in the estimation error, which helps mitigate the impact of sparsity on the overall statistical error.
\end{remark}

\begin{remark} \label{rmk: additional error in mapping}
We acknowledge that, in practice, external semantic information may not perfectly align with the latent embeddings. This mismatch can be represented by the decomposition $\vv_j = \vvarphi(\ve_j) + \vepsilon_j$, where $\vepsilon_j \in \reals^r$ captures the discrepancy between the mapped semantic representation and the latent embedding. Under this extended formulation, the statistical error bound for $\widehat{\mTheta}$ includes an additional approximation error, given by $\mathcal{A}_{n,p,q}^2  = e^{2M_1} \|\mU^{*}\|_{2}^2 (\mu_{q+1}+\|\vepsilon\|_F^2)$ with $\vepsilon = (\vepsilon_1^\top, \cdots, \vepsilon_p^\top)^\top$. When the external embeddings are weakly informative or poorly aligned with the observed matrix data, the term $\|\vepsilon\|_F^2$ can be large. In such cases, the proposed data-driven kernel selection procedure tends to favor the baseline model that does not use external semantic information, thereby avoiding the additional semantic approximation error.
\end{remark}

\subsection{Local Convergence of Projected Gradient Descent Algorithm}
\label{subsec: local convergence}

The constrained optimization problem in~(\ref{eq: global opt}) is non-convex, and finding a global optimum is generally intractable. 
We therefore develop a scalable PGD algorithm to optimize the regularized objective function described in~\Cref{algo: PGD}. 
In this subsection, we further analyze the \emph{local convergence} behavior of the algorithm by characterizing the evolution of its iterates.
In particular, our objective function can be viewed as a convex loss of a low-rank matrix in the form $\mU\mV^\top$, but the  optimization problem over the factored parameters $(\mU,\mV)$ is non-convex. 
For this class of problems, \citet{park2018finding} studied the numerical convergence properties of first-order gradient descent in deterministic settings. 
In contrast, our analysis focuses on how \emph{statistical error} propagates across PGD iterates, and explicitly accounting for sparsity in the binary data matrix and approximation error induced by the KPCA approximation.

The factorization of $\mU \mV_q^{\top}$ into $\mU$ and $\mV_q$ is not unique. We assume that the true parameters satisfy the balancing condition ${\mU^*}^\top \mU^* = {\mV_q^*}^\top \mV_q^*$. Under this condition, their spectral norms satisfy
$
\|\mU^*\|_2^2 = \|\mV_q^*\|_2^2 = \|\mU^* \mV_q^{*\top}\|_2 .
$
Let $(\valpha_t, \rho_t, \mU_t, \mV_t)$ denote the estimates at iteration $t$. 
We define the estimation errors as
$
\dalphat = \valpha_t - \valpha^*, \quad
\drhot = \rho_t - \rho^*, \quad
\dUt = \mU_t - \mU^*\mO_t, \quad
\dVt = \mV_t - \mV_q^*\mO_t,
$
where $\mO_t = \argmin_{\mO \in \mathcal{O}(r)} \|\mU_t - \mU^*\mO\|_F$ aligns the estimated factors with the truth and accounts for the orthogonal invariance of the factorization.
We quantify the estimation error at iteration $t$ using the metric
\[
e_t
= \|\mU^*\|_2^2 \|\dUt\|_F^2
+ \|\mV_q^*\|_2^2 \|\dVt\|_F^2
+ \|\dalphat \vone_p^\top\|_F^2
+ \|\drhot \vone_n \vone_p^\top\|_F^2 .
\]

For theoretical analysis, we further include projection steps to ensure that all iterates remain within the bounded parameter space defined in~(\ref{eq:para space}) in Algorithm~\ref{algo: PGD}.
Specifically, at each iteration we project $\valpha_t$ onto $C_{\alpha} = \{\valpha: \valpha = \mJ_n \valpha, \|\valpha\|_{\infty} \leqslant M\}$, $\mU_t$ onto $C_{u} = \{\mU: \max _{i \in [n]}\left\|\mU_{i*}\right\|^2 \leqslant M\}$, $\mV_t$ onto $C_{v} = \{\mV: \mV = \proj \mV = \mPsi_{[p]}\mGamma_t \text{ with } \|\mGamma\|_F^2 \leqslant M\}$, and $\rho_t$ onto $C_{\rho}=\{\rho: -M_1 \leqslant \rho\leqslant -m_2\}$. 
In practice, we observe that the algorithm performs well even without explicit projection, but these constraints simplify the analysis by guaranteeing the boundedness of the iterates.

We now establish the local convergence properties of the proposed PGD algorithm.
Let $\kappa_{\mU^*\mV_q^{*\top}}$ denote the condition number of the signal matrix $\mU^*\mV_q^{*\top}$. 
Theorem demonstrates that, with a suitable initialization, the estimation error of iterates produced by Algorithm~\ref{algo: PGD} converges linearly it reaches a statistical tolerance floor determined by the estimation and approximation errors. The proofs of~\Cref{thm: algorithm error} and~\Cref{corollary: algorithm error} are provided in~Supplementary Material~C.

\begin{theorem}\label{thm: algorithm error} Suppose Conditions (C1)-(C2) in~\Cref{thm: statistical error} hold, and (C3) the identifiability condition ${\mU^*}^\top \mU^*  = {\mV_q^*}^\top \mV_q^*$ holds; (C4) the initializer $\{\rho_0, \valpha_0 ,\mU_0, \mV_0\}$ from Algorithm~S1 satisfies that $e_0 \leqslant c_0 e^{-2M_1}\|\mU^* \mV_q^{*\top}\|_2^2/\kappa_{\mU^* \mV_q^{*\top}}^2$ for a sufficiently small positive constant $c_0$; and (C5) $\|\mU^* \mV_q^{*\top}\|_2 \geqslant C_0 \kappa_{\mU^* \mV_q^{*\top}}  \cdot \max \{\mc{E}_{n,q}/\sqrt{r}, \sqrt{\eta  e^{M_1}}\mc{E}_{n,q},  \sqrt{\eta }\mc{A}_{n,p,q}\}$ with a sufficiently large constant $C_0 >0$,. Then, there exist constants $\eta$, $c$, and $C$ such that uniformly over $ \mathcal{F}\left(n, p, q, M_1, M_2\right)$, the output of the $t$-th iteration in Algorithm~\ref{algo: PGD} satisfies, with probability at least $1-(n+q)^{-c}$, 
\[e_t \leqslant 2\left( 1- \frac{\tau}{e^{M_1 }\kappa_{\mU^* \mV_q^{*\top}}}\eta \right)^t e_0 + \frac{C \kappa_{\mU^* \mV_q^{*\top}}}{\eta }\left( \mathcal{E}_{n,q}^2 + \mathcal{A}_{n,p,q}^2 \right).\]
It follows that for any $T \geqslant T_0 :=  \log(\frac{(n+p)^2M^2}{e^{3M_1-M_2}\kappa_{\mU^* \mV_q^{*\top}}^2})\bigg/\log \left((1-\frac{\eta \nu }{e^{M_1}\kappa_{\mU^* \mV_q^{*\top}}})^{-1}\right)$, 
\[e_T\leqslant C' \kappa_{\mU^* \mV_q^{*\top}} \left( \mathcal{E}_{n,q}^2 + \mathcal{A}_{n,p,q}^2 \right),\]
holds with some constant $C'$.
\end{theorem}

The first term in the high-probability error bound in~\Cref{thm: algorithm error} corresponds to the optimization error arising from numerical iterations. 
This term decays geometrically and vanishes as the number of iterations increases. 
After the algorithm is run for more than $T_0$ iterations, 
the error is dominated by the statistical error in the second term.

We further extend this result to the estimation error of the full matrix parameter $\mTheta$. Define the estimator at step $t$ as $\mTheta_t = \rho_t \vone_n \vone_p^\top + \valpha_t \vone_p^\top + \mU_t \mV_t^\top$. 
The following corollary shows that the estimated matrix achieves the same statistical error as in Theorem~\ref{thm: statistical error}, up to a multiplicative factor $\kappa_{\mU^* \mV_q^{*\top}}$.

\begin{corollary}\label{corollary: algorithm error}
	Assume Conditions (C1)-(C5) in \Cref{thm: algorithm error} hold, then there exist constants $\eta$, $c$, and $C$ such that uniformly over $ \mathcal{F}\left(n, p, q, M_1, M_2\right)$, the output of Algorithm~\ref{algo: PGD} after $T\geqslant T_0$ iterations satisfies, with probability at least $1-(n+q)^{-c}$, 
\[\|\mTheta_T- \mTheta^*\|_F^2 \leqslant C \kappa_{\mU^* \mV_q^{*\top}} \left( \mathcal{E}_{n,q}^2 + \mathcal{A}_{n,p,q}^2 \right).\]
\end{corollary}

\section{Simulation Studies}\label{sec: simulation}
In this section, we conduct extensive simulation studies to assess the impact of three factors on the estimation error of KELP in the imbalanced regime: the number of rows~$n$, the number of columns~$p$, and the sparsity level of the binary matrix. We also investigate the effectiveness of data-driven kernel selection procedure under various choices of the link function~$\vvarphi$ in~Supplementary Material~F, and find that it consistently identifies the best-performing candidate.

\paragraph{Methods for Comparison.} We compare our proposed method with a standard generalized linear factor model that does not use external embeddings. The baseline method relies solely on the data itself and estimates parameters by optimizing (\ref{eq: nonconvex obj}) without the constraint $\mV \in col({\mPsi}_{[q]})$. 
In contrast, KELP leverages external embeddings for knowledge fusion and requires selecting a kernel. We consider four candidate kernels: a linear kernel $\mc{K}(\ve_1, \ve_2) = \ve_1^\top \ve_2$ and three Gaussian kernels $\mc{K}(\ve_1, \ve_2;\mb{r}) =\exp(-\|\ve_1- \ve_2\|_2^2 /(2\mb{r}^2)) $ with radius parameters $1/(2\mb{r}^2) = 0.001, 0.01, 0.1$. For each candidate kernel, we select the dimension of the kernel PCA  to be the smallest $q$ such that $\sum_{\ell=1}^q \tau_\ell / \sum_{\ell=1}^p \tau_\ell \geq 0.95$, where $\left\{\tau_j: j \in [p]\right\}$ are the non-decreasing eigenvalues of $\mK_c$. 
The data-driven kernel selection approach introduced in Section~\ref{sec: data-driven} is applied for kernel selection.

\paragraph{Simulation Settings.} We generate data following  model~(\ref{eq: glfm-kf}) under both linear and nonlinear mappings~$\vvarphi$. 
For a given number of rows $n$, we generate row entity embeddings $\vu_i \overset{iid}{\sim} \mathcal{N}(\vzero_r, \mI_r)$ with latent dimension $r=8$ for $i \in [n]$. We center the columns of $\mU$ by setting $\mU \leftarrow \mJ_n \mU$, where $\mJ_n = \mI_n - \frac{1}{n}\vone_n \vone_n^\top $. 
For a given number of columns $p$, we generate  semantic embeddings $\{\ve_j\}_{j=1}^p$ with a clustering structure that reflects patterns commonly observed in semantic representations.
Specifically, we sample $K=10$ cluster centers $\{{\vc_k}\}_{k=1}^K$ uniformly from the unit sphere in $d=50$ dimensional space to represent topics and then randomly assign each column entity $j$ to a cluster $z_j \in [K]$. We generate a perturbed vector $\tilde{\ve}_j = \vc_{z_j}+0.05\epsilon_j$ with $\epsilon_j \overset{iid}{\sim} \mathcal{N}(\vzero_d, \mI_d)$ and normalize them to obtain $\ve_j = {\tilde{\ve}_j}/{\|\tilde{\ve}_j\|_2}$. 
The column entity embeddings $\vv_j = [\varphi_1(\ve_j), \cdots, \varphi_r(\ve_j)]^\top$ are derived from the semantic embeddings via two types of mapping functions: 
\begin{itemize}
	\item[(1)] Linear function: $\varphi_k(\ve_j) = \vw_k^\top  \ve_j$ with coefficients $\vw_k \overset{iid}{\sim} \mathcal{N}(\vzero_d, 2\mI_d)$ for $k \in [r]$. 
	\item[(2)] Smooth nonlinear function: $\varphi_k(\ve_j) = \tanh(\vw_{2k}^\top (\vw_{1k}^\top \ve_j)^2)$ with coefficients $\vw_{1k} \overset{iid}{\sim} \mathcal{N}(\vzero_{d\times 2r}, \mI_{d\times 2r})$ and $\vw_{1k} \overset{iid}{\sim} \mathcal{N}(\vzero_{2r \times r}, \mI_{2r \times r})$ for $k \in [r]$.
\end{itemize}
After generating the column embeddings $\mV = [\vv_1^\top , \cdots, \vv_p^\top ]^\top $, we center the columns by $\mV = \mJ_p \mV$ with $\mJ_p = \mI_p - \frac{1}{p}\vone_p \vone_p^\top $. To ensure identifiability, we perform the singular value decomposition of $\mU \mV^\top = \widetilde{\mU}\mD \widetilde{\mV}^\top$ and define $\mU^* = \widetilde{\mU}\mD^{1/2}$ and $\mV^* = \widetilde{\mV}\mD^{1/2}$. We further normalize $\mU^*$ and $\mV^*$ elementwise such that $\|\mU^*\mV^{*\top}\|_F^2 = np$. 
The subject heterogeneity parameters are generated as $\alpha_i \overset{iid}{\sim} U(-1,1)$ for $i \in [n]$ and centered to obtain $\valpha^* = \mJ_n \valpha$. 
The sparsity parameter is fixed at $\rho^* = -1.5$ (yielding approximately 23\% non-zero entries), unless we assess the impact of sparsity specifically.  
Given the true parameter matrix $\mTheta^* = \rho^* \vone_n \vone_p^\top + \valpha^* \vone_p^\top + \mU^* \mV^{*\top}$, we generate 20 replications of a binary matrix $\mY$.
We evaluate the relative estimation error for $\mTheta$, defined by $\|\widehat{\mTheta} - \mTheta^*\|_F / \|\mTheta^*\|_F$.

\begin{figure}[!h]
\begin{subfigure}{.33\textwidth}
  \centering
  \includegraphics[width=\linewidth]{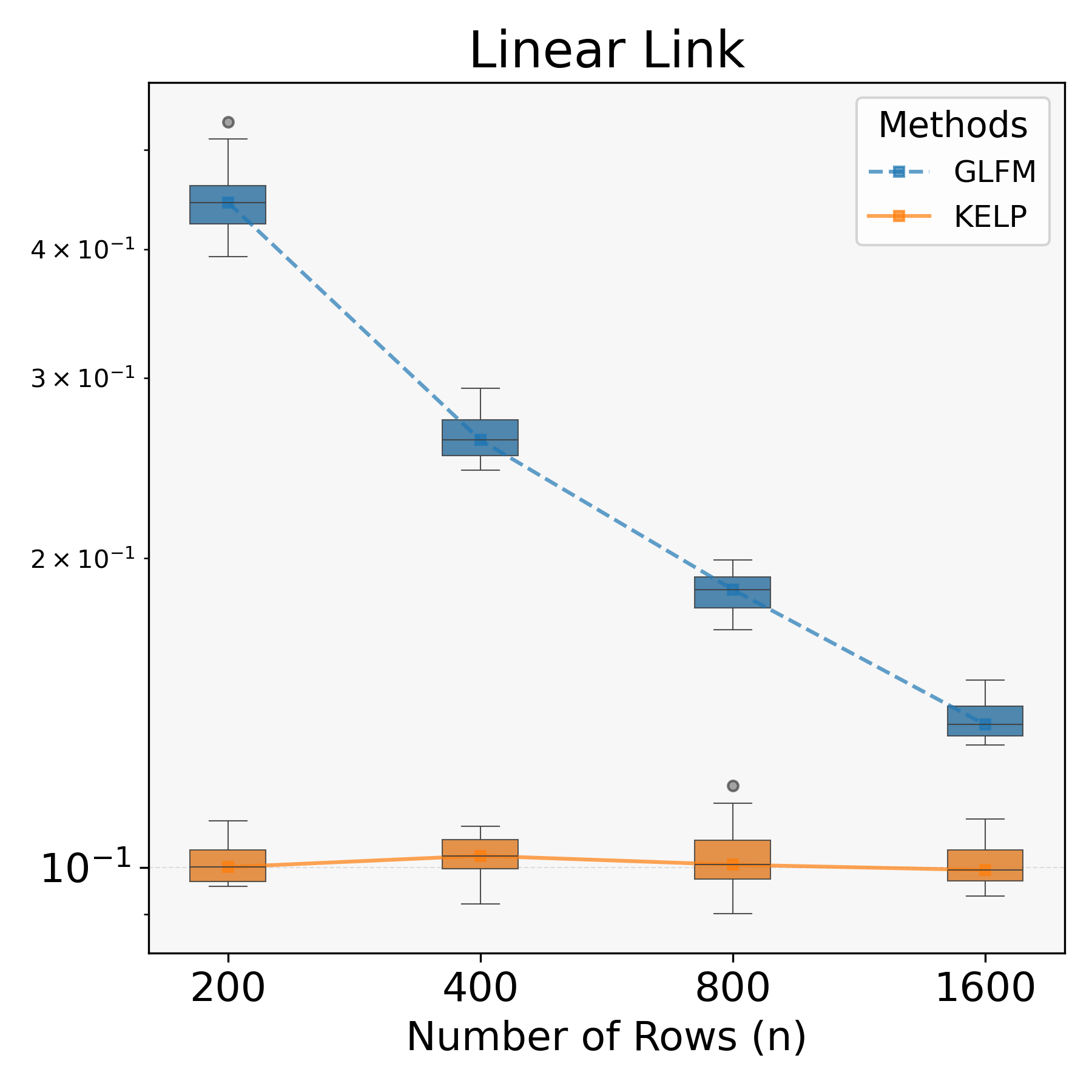}
  \includegraphics[width=\linewidth]{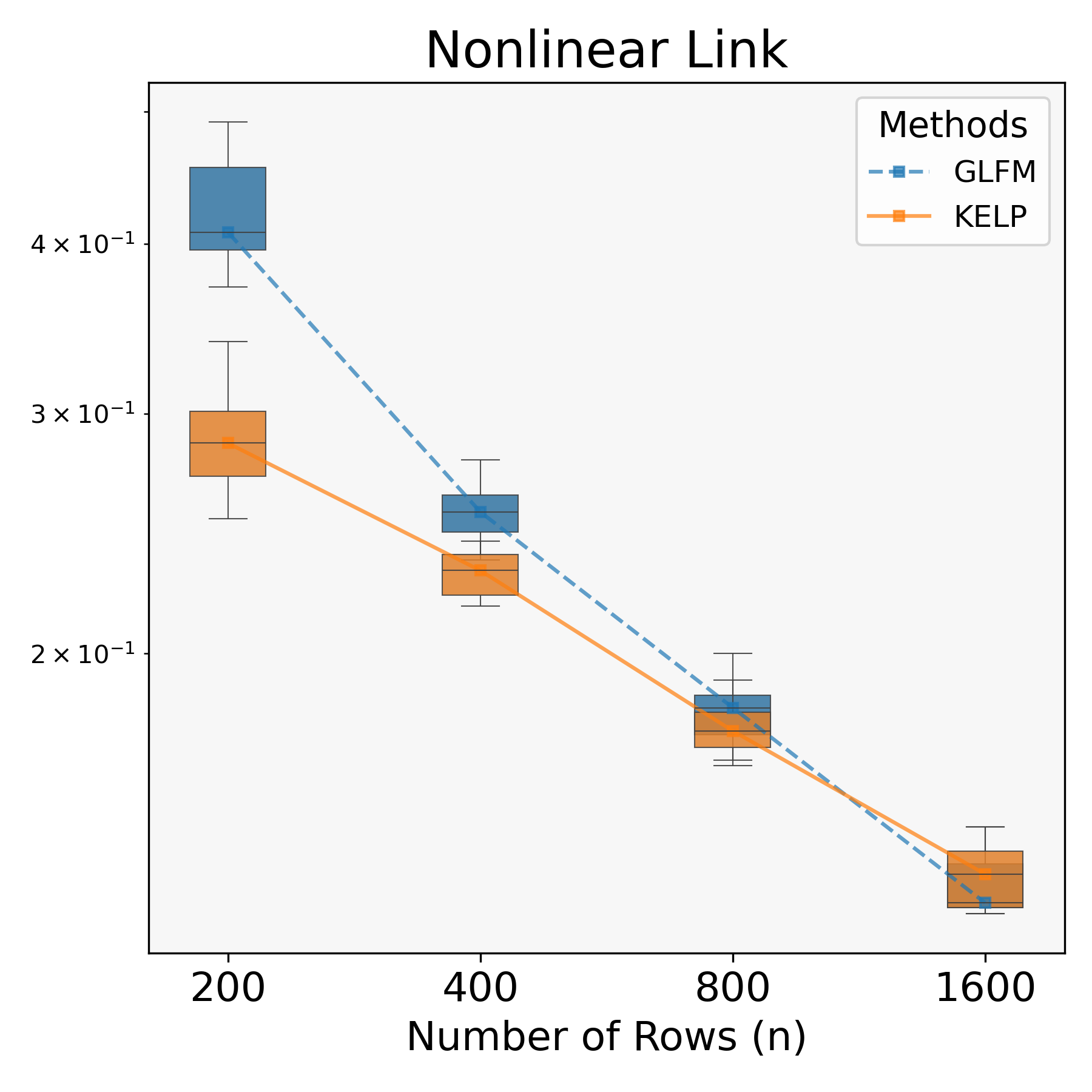}
  \caption{Varying $n$}
  \label{fig:varying n}
\end{subfigure}%
\begin{subfigure}{.33\textwidth}
  \centering
  \includegraphics[width=\linewidth]{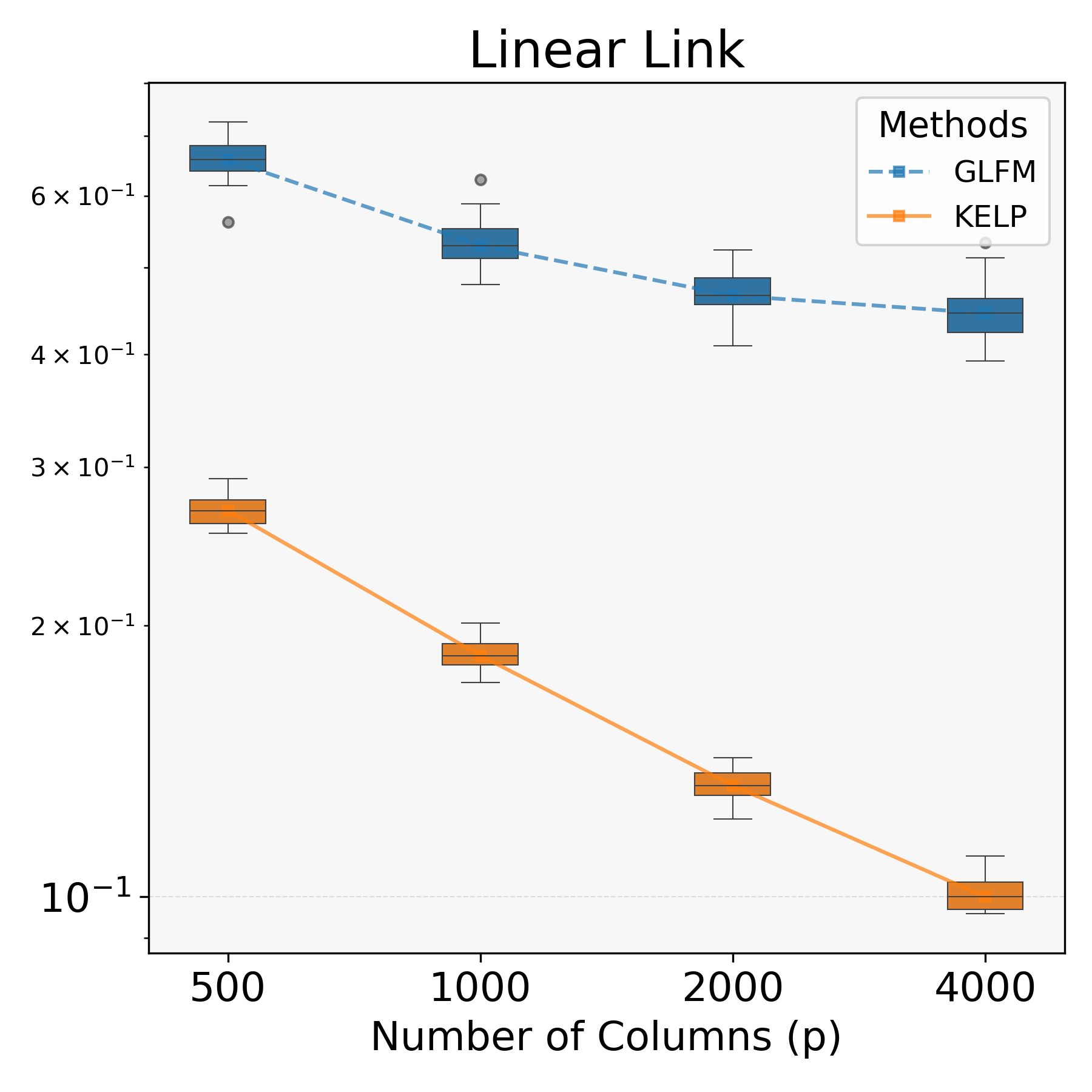}
  \includegraphics[width=\linewidth]{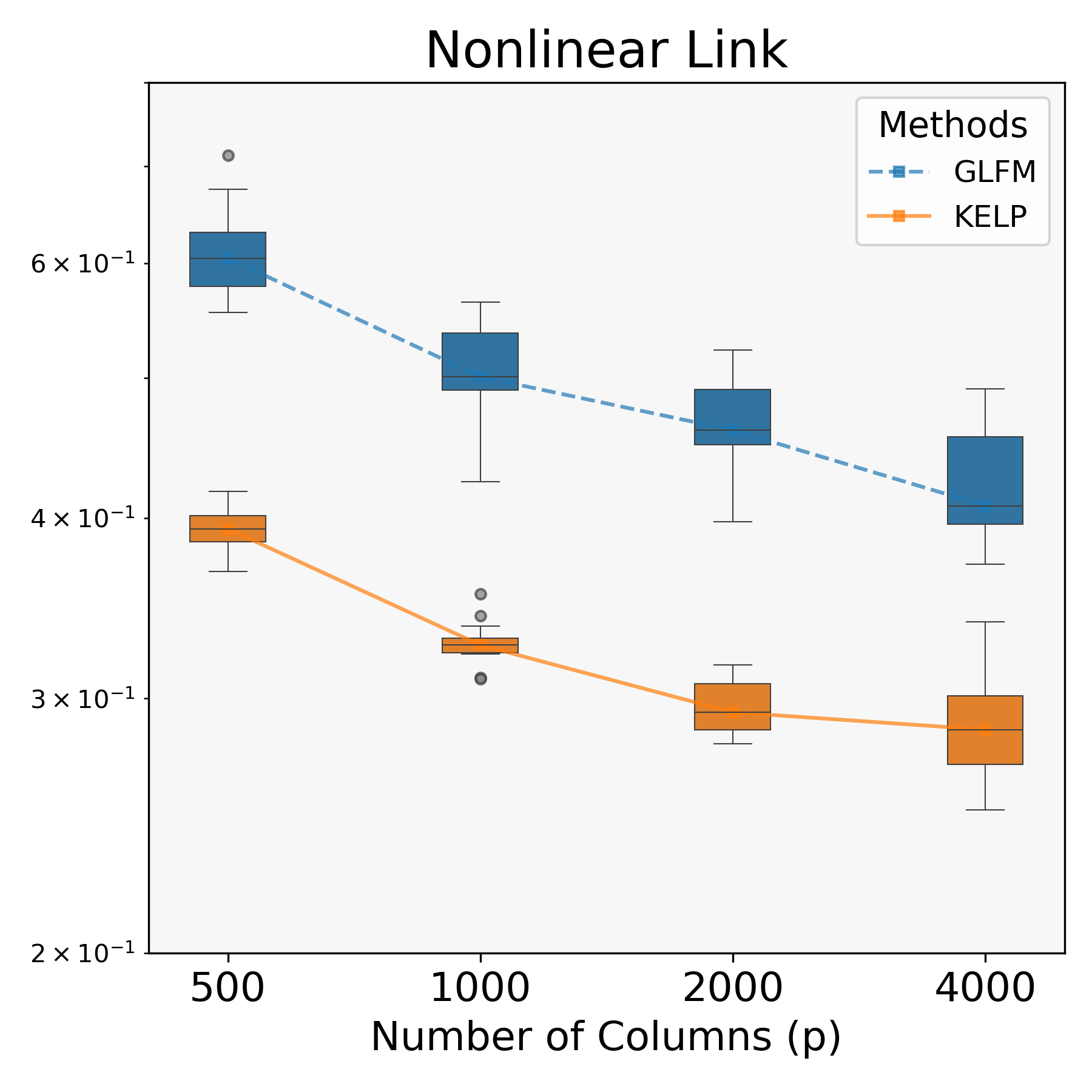}
  \caption{Varying $p$}
  \label{fig:varying p}
\end{subfigure}
\begin{subfigure}{.33\textwidth}
  \centering
  \includegraphics[width=\linewidth]{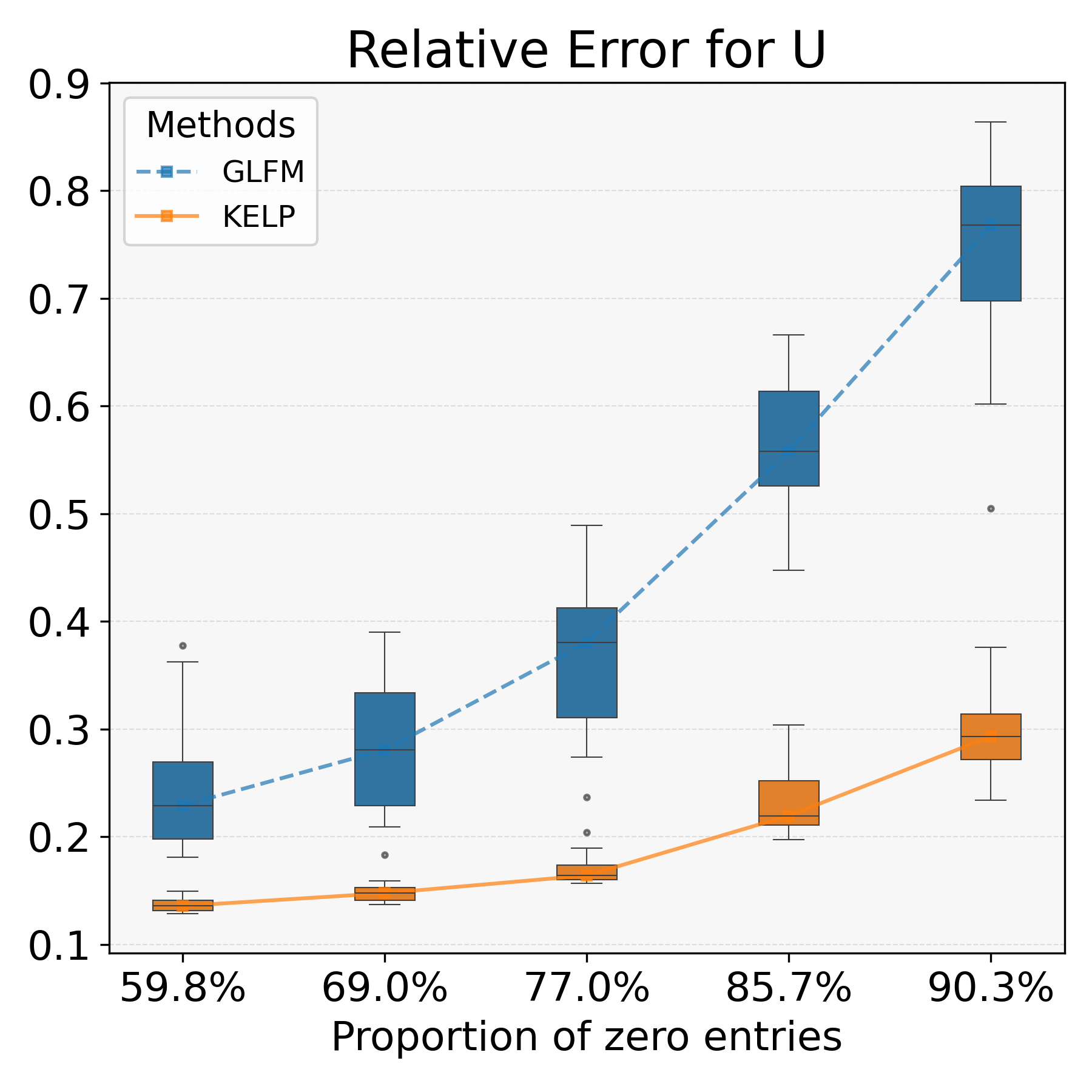}
  \includegraphics[width=\linewidth]{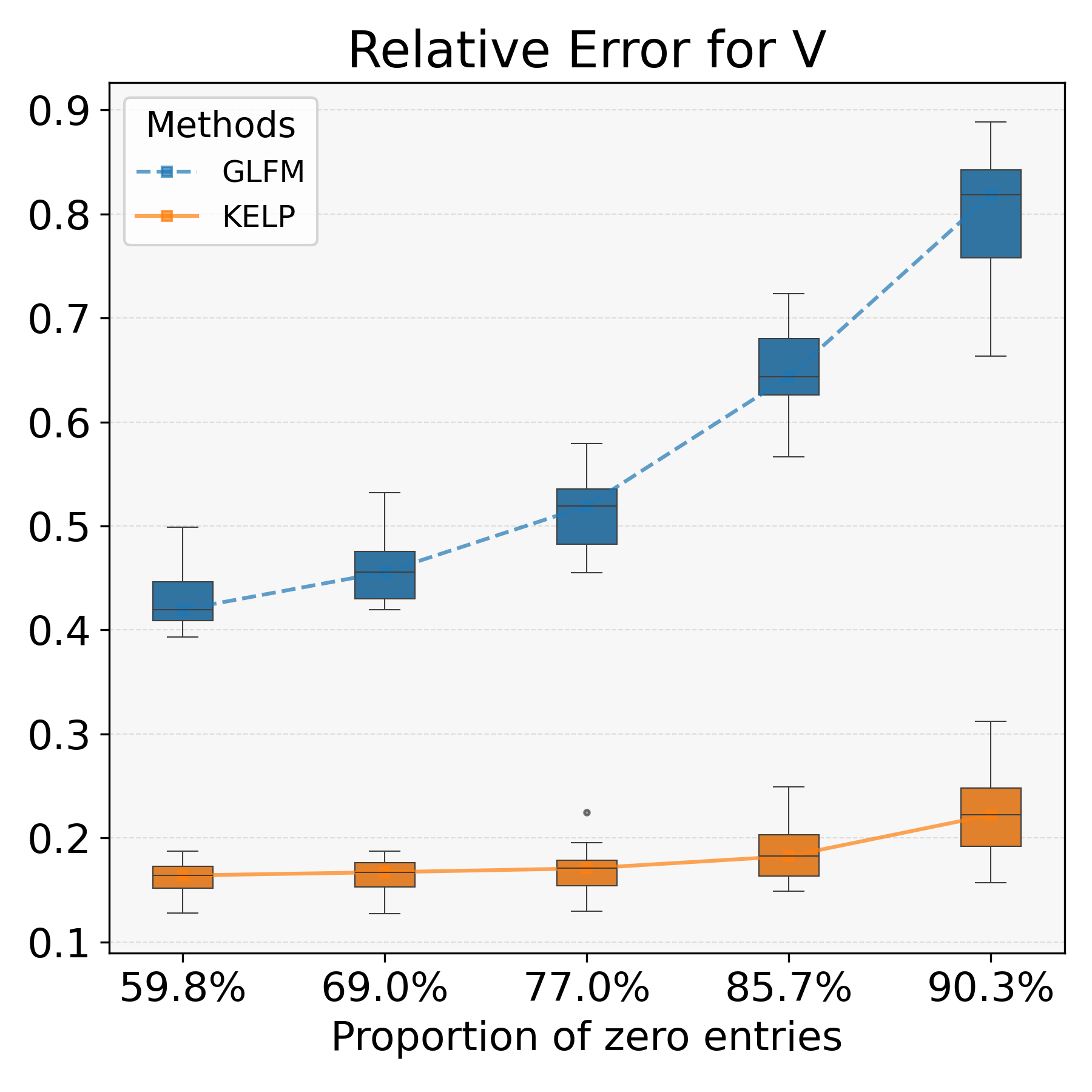}
  \caption{Varying Sparsity}
  \label{fig:varying rho}
\end{subfigure}
\caption{Comparison between the proposed KELP estimator with data-driven kernel selection and the standard GLFM. \textbf{Panels (a) and (b)} show log-log plots of the relative estimation error for $\mTheta$ versus the number of rows $n$ and the the number of columns $p$, respectively. The upper and lower rows correspond to the linear (1) and nonlinear (2) mapping function settings. \textbf{Panel (c)} shows the relative errors for $\mU$ and $\mV$ versus the proportion of zero entries (sparsity level) under the linear mapping setting.}
\end{figure}

\subsection{Varying Sample Size}
We fix the number of columns at $p=4000$ and vary $n$ from $200$ to $1600$. 
As shown in~\Cref{fig:varying n}, our method consistently achieves lower or comparable errors than the baseline in this imbalanced regime ($n \ll p$).
In the linear mapping setting, where the linear kernel is correctly specified and misspecification error is negligible, our method yields smaller errors across all sample sizes, while the advantage diminishes as $n$ grows. 
In the nonlinear mapping setting, when $n$ is small, estimation error dominates and our method performs better by effectively reducing dimensionality via leveraging external semantic embeddings. As $n$ grows, the approximation error brought by mis-specified kernel becomes more salient. The data-driven kernel selection procedure then chooses the baseline to avoid negative knowledge fusion, which results in the same errors for both methods. 
We also observe that the relative errors for the baseline decrease at the rate of $n^{-1/2}$ for large $n$ in both settings, while our method stabilizes in the linear mapping setting.  This aligns with our theoretical results. Note that the estimation error for the baseline, $\sqrt{(n+p)/(np)}$, reduces to $n^{-1/2}$ when $n \ll p$, and that for our method $\sqrt{(n+q)/(np)}$ becomes $p^{-1/2}$ in the correctly specified linear setting ($q = d = 50 \ll n $), which remains constant as $n$ increases.

\subsection{Varying Number of Features}
We fix the sample size at $n = 200$ and vary the number of columns $p$ from $500$ to $4000$. 
As shown in~\Cref{fig:varying p}, the relative errors for both methods decrease as $p$ increases. In all cases, KELP achieves smaller errors than the baseline method. 
Moreover, in the linear mapping setting, the rate of error reduction with respect to $p$ is faster for KELP compared to the baseline. This observation also aligns with our theory. When $n$ is fixed and $p$ grows, the estimation error bound for the baseline method approaches a constant, as $\sqrt{(n+p)/(np)} \asymp n^{-1/2}$ for large $p$. 
In contrast, under a correctly specified linear kernel, the estimation error for KELP is $\sqrt{(n+q)/(np)} \asymp p^{-1/2}$, as $q=d=50$ and $n$ are fixed and do not grow with~$p$.

\subsection{Varying Sparsity of Feature Frequency} \label{subsec: simu sparsity}
We fix $n=400$ and $p=4000$ and vary $\rho$ such that the proportion of zero entries in the observation matrix ranges from 60\% to 90\%. 
In this experiment, we report the relative errors for $\mU$ and $\mV$ instead of $\mTheta$. This is because the change of $\rho$ will affect the scale of $\mTheta^*$, which makes it tricky to compare the relative errors for $\mTheta$ across different values of~$\rho$. In contrast, the scales of $\mU^*$ and $\mV^*$ are not affected by $\rho$. The relative error for $\mU$ (and similarly for $\mV$) is defined as $\min_{\mO \in \mathcal{O}(r)} \|\widehat{\mU} - \mU^*\mO\|_F / \|\mU^*\|_F$ to account for the identifiability up to the rotation.
As shown in~\Cref{fig:varying rho}, the relative errors for both methods tend to increase as the binary matrix becomes sparser. However, the baseline suffers more from high sparsity level, as indicated by the significant increase of error for the baseline method when the proportion of zero entries increases from $77.0\%$ to $90.3\%$. For KELP, we also observe that the error for $\mV$ is less sensitive to the sparsity level compared to that for $\mU$. This is because the estimate of $\mV$ is directly enhanced by the external semantic embeddings, while the estimate of $\mU$ benefits from the external knowledge indirectly through the improved estimate of $\mV$ within the iterative PGD algorithm.

\section{Application to Electronic Health Records}
\label{sec: real data}

In this section, we apply KELP to analyze real-world EHR data. 
Patient-level EHR data can be represented as a matrix, with patients as rows and clinical features as columns. 
Latent space models provide a useful tool for learning patient and feature embeddings. Such embeddings can support various downstream tasks such as knowledge graph construction and risk prediction~\citep{choi2018mime,hong2021clinical, gan2025arch}.
Our study focuses on multiple sclerosis (MS), an autoimmune neurological disease associated with long-term disability.
While many efforts have been made to produce general-purpose semantic embeddings for EHR features from broad patient cohorts in large health systems \citep{hong2021clinical, gan2025arch}, such embeddings may be too coarse to capture MS-specific clinical patterns.
Learning specialized embeddings from MS cohorts alone, however, is challenging. 
Because MS diagnosis typically requires manual confirmation via chart review of clinical history, unstructured notes, and imaging, which results in specialized MS cohorts with limited sample size. 
Combined with the high dimensionality of EHR features, this creates an imbalanced regime in which standard GLFM struggles to achieve reliable parameter estimation. 
To address this, we leverage existing semantic embeddings extracted from a large general cohort to improve latent space modeling in a small MS cohort. 

\subsection{Cohort Description and Data Processing}
We analyze data from patients enrolled in a clinic-based MS cohort, {\em Prospective Investigation of Multiple Sclerosis in the Three Rivers Region} (PROMOTE,
Pittsburgh, PA), between 2017 and 2023. 
We included $n=212$ patients with a confirmed diagnosis of MS, validated via manual chart review, and required at least 24 months of disease history between diagnosis and enrollment. 
For each patient, we use their EHR data from the University of Pittsburgh Medical Center (UPMC) within a 12-month window centered on their enrollment date (i.e., from 12 months before to 12 months after enrollment). We aggregate all clinical events observed within this window into binary indicators of feature occurrence. The resulting dataset is a patient-by-feature binary matrix, where each row corresponds to a patient and each column indicates the presence or absence of a clinical concept during the aggregation window. 
The features span multiple domains, including \textit{PheCode} codes for diagnoses, \textit{clinical classification software (CCS)} codes for procedures and services~\citep{wei2017evaluating}, \textit{RxNorm} codes for medication prescriptions~\citep{nelson2011normalized}, and \textit{concepts unique identifiers (CUIs)} extracted from free-text clinical narratives using a validated natural language processing pipeline according to unified medical language system (UMLS)~\citep{yu2015toward}. 
The final processed dataset contains $p=3,296$ features, including $327$ PheCodes, $72$ CCS codes, $234$ RxNorm codes, and $2,663$ CUIs. 
Approximately 88.3\% of the matrix entries are zero.
As a source of external knowledge, we use existing clinical semantic embeddings (128 dimensions) constructed from the EHR data of 12.5 million Veterans Affairs (VA) patients \citep{gan2025arch}. 
These high-quality embeddings have been validated across a wide range of generic clinical tasks such as identifying drug side effects and disease phenotyping.

\subsection{Validation Results}
In this imbalanced regime, we first use a matrix completion task to guide the choice of latent dimensionality for the MS cohort, and then evaluate our knowledge-fused latent space model through two downstream tasks: MS knowledge graph reconstruction and patient disability phenotyping.
To identify an appropriate range of ranks, we randomly mask 20\% of the matrix entries for evaluation and fit the model on the remaining 80\%. We compare our knowledge-fused model with a standard GLFM baseline, and report the Area Under the Receiver Operating Characteristic curve (AuROC) based on estimated probabilities over $10$ random splits and varying latent ranks. As shown in~Figure~S2 in Supplementary Material, AuROC for our model increases through moderate ranks and reaches its peak at $r=8$. The knowledge-fused model also achieves higher AuROC than the GLFM at this rank, which indicates that leveraging external semantic embeddings improves capturing the data structure. 
Importantly, these results suggest that the MS cohort can be effectively represented in a relatively low-dimensional latent space. 
Whereas large-scale general-purpose medical embeddings (e.g., VA embeddings trained on 12.5 million patients) require high dimensionality due to broader population heterogeneity, the MS cohort forms a more specialized and semantically coherent sub-population, allowing its structure to be captured with substantially fewer latent dimensions.
Guided by this validation, we focus on ranks $r \in \{2, 4, 8, 16\}$ for subsequent analyses.  In the following, we assess the utility of the learned embeddings in the low-dimensional latent space for MS knowledge reconstruction and disability phenotyping.

\begin{figure}[!h]
\begin{subfigure}{.33\textwidth}
  \centering
  \includegraphics[width=\linewidth]{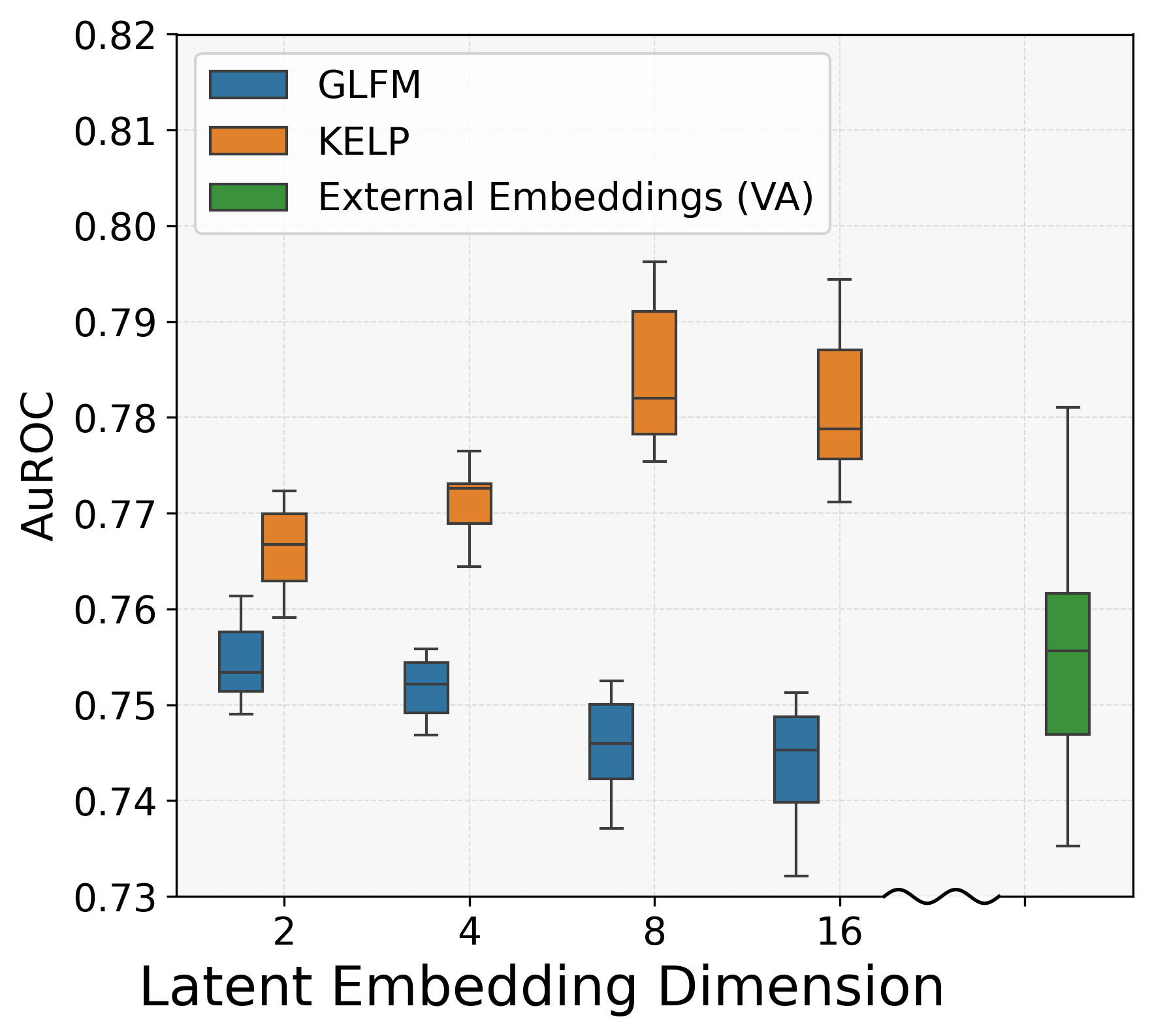}
  \caption{Knowledge Graph Curation}
  \label{fig:kg}
\end{subfigure}%
\begin{subfigure}{.67\textwidth}
  \centering
  \includegraphics[width=\linewidth]{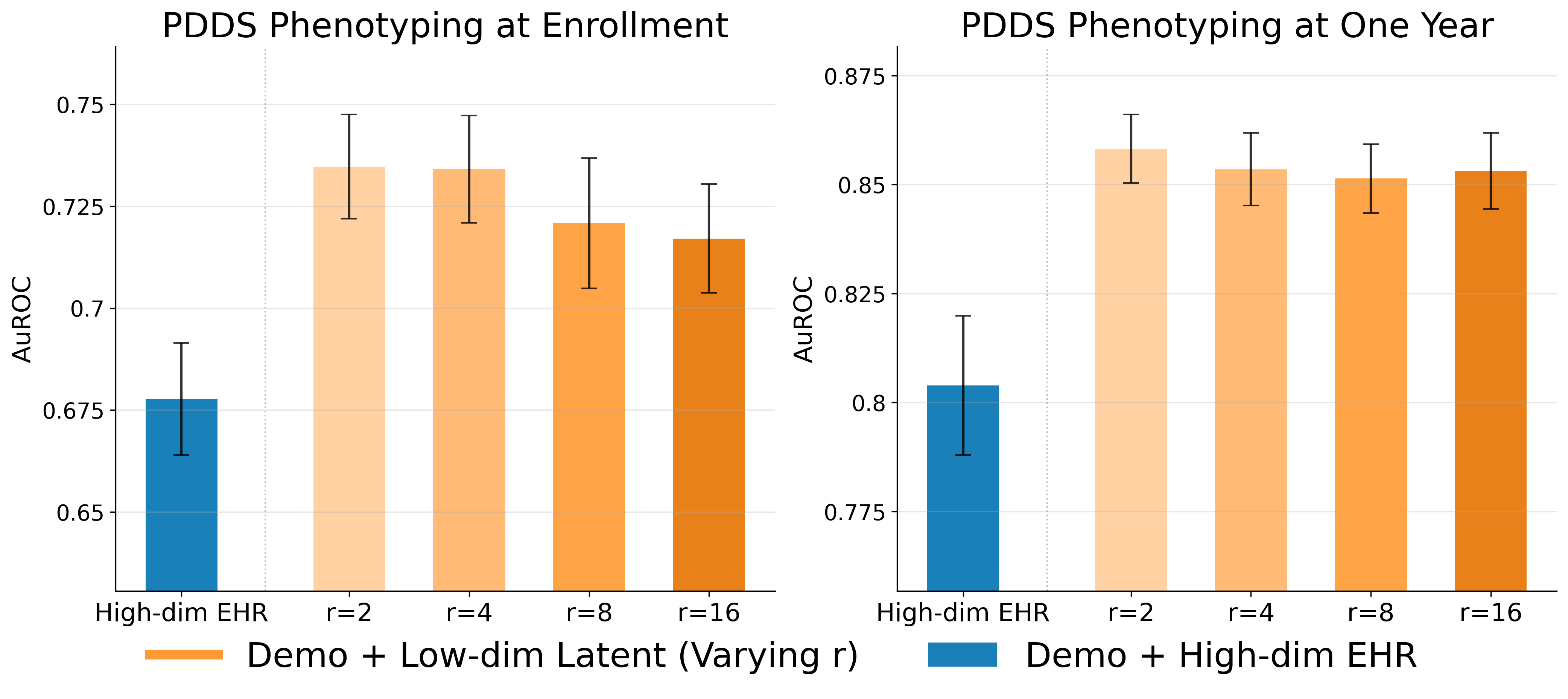}
  \caption{Disability Phenotyping}
  \label{fig:pdds}
\end{subfigure}
\caption{Evaluation of learned feature and patients embeddings on clinical downstream tasks. \textbf{Panel (a)} reports AuROC for recovering known MS-related clinical relationships, comparing three sets of feature embeddings: knowledge-fused LSM, GLFM trained solely on the MS cohort, and the original VA embeddings. \textbf{Panel (b)} reports AuROC for severe disability phenotyping at enrollment (left) and one-year follow-up (right), comparing two feature sets: the patient embeddings in an $r$-dimensional latent space and the original high-dimensional binary EHR features. 
}
\end{figure}

\paragraph{MS Knowledge Reconstruction.} 
We evaluate the quality of the learned feature embeddings by testing their ability to recover known MS-related clinical relationships. 
A curated set of related feature pairs was extracted from the UMLS~\citep{bodenreider2004unified} and a validated online narrative and codified feature search engine \citep{xiong2023knowledge}. A pair is labeled as MS-related if at least one element corresponds to MS PheCode or MS CUI, resulting in 96 known positive pairs.  
To construct negative controls, we randomly sample feature pairs that match the semantic type of the positive group (e.g., disease–drug pairs when assessing ``may treat or may prevent'' relations), which is a common strategy adopted in validating knowledge graph~\citep{hong2021clinical, gan2025arch}.
Due to the limited number of positives, we repeat the negative sampling $30$ times. This produces 30 balanced evaluation sets.
Heuristically, high-quality features embeddings should place clinically associated concepts close together in latent space.
We quantify pairwise relatedness using the inner product between embedding vectors, and evaluate discrimination using the AuROC, which measures the probability that a known MS-related pair attains a higher similarity score than a randomly paired control.
We compare three feature embeddings: (1) our knowledge-fused embeddings, (2) embeddings from a standard GLFM model trained solely on the MS cohort, and (3) the original 128-dimensional VA embeddings.
As shown in \Cref{fig:kg}, our method achieves higher AUROC than both baselines. Importantly, it outperforms the original VA embeddings, indicating that integrating general knowledge with MS-specific cohort data refines the latent structure and brings MS-relevant concepts into closer proximity. This improvement is consistent across ranks, with the best performance observed at the selected optimal rank $r=8$.

\paragraph{Disability Phenotyping.}
Finally, we evaluate the clinical utility of the patient embeddings in the latent space through disability phenotyping tasks. 
We measure disability using the patient determined disease steps (PDDS) scale, a patient-reported outcome ranging from $0$ to $8$, with higher scores indicating greater functional impairment. 
We focus on the clinically meaningful threshold PDDS $\ge 4$, which corresponds to the onset of mobility limitation (e.g., intermittent or constant walking assistance)~\citep{marrie2006does}.
We consider two related phenotyping tasks. First, we identify disability status at enrollment, where the enrollment PDDS score is defined using measurements collected within a window of $\pm 1$ month around the enrollment date. Second, we phenotype short-term disability status at approximately one year after enrollment, where the one-year PDDS is defined using measurements collected between months 11 and 13. For both tasks, EHR features are derived from the same aggregation window centered on enrollment.
We compare two sets of predictors in a logistic regression model with a Lasso penalty: (1) the original binary occurrences of high-dimensional EHR features; (2) the low-dimensional patient embeddings derived from our model. 
All models include demographic features and disease duration as predictors. 
In the one-year disability prediction task, we additionally adjust for baseline PDDS to account for disease trajectory.
We randomly sample 80\% as training data and assess performance on the remaining 20\% held-out set using AuROC. Given the small sample size with labeled outcomes, we perform $30$ replications of data splitting. 
As shown in~\Cref{fig:pdds}, the low-dimensional embeddings consistently outperform the high-dimensional raw EHR features across embedding ranks in both phenotyping tasks. 
While the inclusion of baseline PDDS leads to higher absolute scores in the one-year follow-up task, the advantage of using low-dimensional embeddings remains consistent.
This indicates that KELP effectively transforms high-dimensional sparse EHR data into informative low-dimensional dense representations suitable for downstream tasks with limited sample~size.

\section{Discussion}

In this work, we have introduced a KELP model that  incorporates available external semantic embeddings to address estimation challenges in high-dimensional, imbalanced, and sparse binary matrices. 
By modeling column embeddings as smooth functions of external semantic information within an RKHS framework, KELP effectively reduced the dimensionality of the latent space and achieves improved estimation error rates compared to standard generalized latent factor models in the $n \ll p$ regime. 
We further proposed a projected gradient descent algorithm, which converges linearly to the statistical error. 
Empirical results on a MS cohort demonstrate that fusing general biomedical knowledge with small-sample clinical data enhances latent structure recovery, leading to improved knowledge graph reconstruction and patient disability risk prediction.

In many applications, additional subject-level covariates are available, such as demographic characteristics or measures of healthcare utilization in EHR. The proposed KELP model can be extended to incorporate subject-level covariates $\{\vx_i\}_{i\in[n]} \subset \reals^m$ by specifying, for $i\in[n]$ and $j\in[p]$, $\mathcal{P}(y_{ij}=1 \mid \{\vx_i\}_{i\in[n]},\{\ve_j\}_{j \in [p]}) =\sigma(\rho+ \alpha_i + \vu_i^\top \vvarphi(\ve_j) + \vx_i^\top \vbeta)$, where $\vbeta\in \reals^m$ denotes the coefficient vector. Our estimation procedure can be naturally extended through a post-estimation decomposition of  $\hat{\valpha}$, which helps separate covariate-driven effects from latent components that are not explained by observed covariates. and latent components that cannot be explained by covariates. More flexible formulations with entity-specific covariate effects would introduce additional identifiability challenges and require new structural assumptions and estimation strategies. We leave a comprehensive investigation of covariate-adjusted KELP models to future work.

There are other promising future directions. First, while our analysis focuses on EHR data aggregated over a fixed time window, patient health is inherently dynamic. Extending the framework to longitudinal settings would allow latent patient representations to evolve over time, enabling the study of disease progression and time-varying risk profiles.
Second, the framework could be broadly applicable to other domains with high-dimensional imbalanced discrete data matrices.
For example, in the evaluation of artificial intelligence systems, performance data are often organized as model–task matrices, where the number of tasks far exceeds the number of models. Incorporating task description embeddings or metadata as semantic side information could enable more robust leaderboard ranking and performance prediction in data-scarce benchmarking settings.

\spacingset{1.5} 
\bibliographystyle{chicago}
\bibliography{reference}

@book{lavravc2021representation,
	author = {Lavra{\v{c}}, Nada and Podpe{\v{c}}an, Vid and Robnik-{\v{S}}ikonja, Marko},
	publisher = {Springer},
	title = {Representation learning},
	year = {2021}}

@article{kopf2021latent,
	author = {Kopf, Andreas and Claassen, Manfred},
	journal = {Patterns},
	number = {3},
	publisher = {Elsevier},
	title = {Latent representation learning in biology and translational medicine},
	volume = {2},
	year = {2021}}

@article{van2017neural,
	author = {Van Den Oord, Aaron and Vinyals, Oriol and others},
	journal = {Advances in neural information processing systems},
	title = {Neural discrete representation learning},
	volume = {30},
	year = {2017}}

@article{hoff2002latent,
	author = {Hoff, Peter D and Raftery, Adrian E and Handcock, Mark S},
	journal = {Journal of the American Statistical Association},
	number = {460},
	pages = {1090--1098},
	publisher = {Taylor \& Francis},
	title = {Latent space approaches to social network analysis},
	volume = {97},
	year = {2002}}

@article{li2020inferring,
	author = {Li, Yue and Nair, Pratheeksha and Lu, Xing Han and Wen, Zhi and Wang, Yuening and Dehaghi, Amir Ardalan Kalantari and Miao, Yan and Liu, Weiqi and Ordog, Tamas and Biernacka, Joanna M and others},
	journal = {Nature Communications},
	number = {1},
	pages = {2536},
	publisher = {Nature Publishing Group UK London},
	title = {Inferring multimodal latent topics from electronic health records},
	volume = {11},
	year = {2020}}

@article{ma2023generalized,
	author = {Ma, Ting Fung and Wang, Fangfang and Zhu, Jun},
	journal = {Biometrics},
	number = {3},
	pages = {2311--2320},
	publisher = {Wiley Online Library},
	title = {On generalized latent factor modeling and inference for high-dimensional binomial data},
	volume = {79},
	year = {2023}}

@article{liu2024representation,
	author = {Liu, Suqi and Cai, Tianxi and Li, Xiaoou},
	journal = {arXiv preprint arXiv:2410.07454},
	title = {Representation-Enhanced Neural Knowledge Integration with Application to Large-Scale Medical Ontology Learning},
	year = {2024}}

@article{ma2020universal,
	author = {Ma, Zhuang and Ma, Zongming and Yuan, Hongsong},
	journal = {Journal of Machine Learning Research},
	number = {4},
	pages = {1--67},
	title = {Universal latent space model fitting for large networks with edge covariates},
	volume = {21},
	year = {2020}}

@article{marrie2006does,
	author = {Marrie, RA and Cutter, G and Tyry, T and Vollmer, T and Campagnolo, D},
	journal = {Neurology},
	number = {8},
	pages = {1235--1240},
	publisher = {Lippincott Williams \& Wilkins},
	title = {Does multiple sclerosis--associated disability differ between races?},
	volume = {66},
	year = {2006}}

@article{nelson2011normalized,
	author = {Nelson, Stuart J and Zeng, Kelly and Kilbourne, John and Powell, Tammy and Moore, Robin},
	journal = {Journal of the American Medical Informatics Association},
	number = {4},
	pages = {441--448},
	publisher = {BMJ Group BMA House, Tavistock Square, London, WC1H 9JR},
	title = {Normalized names for clinical drugs: RxNorm at 6 years},
	volume = {18},
	year = {2011}}

@article{wei2017evaluating,
	author = {Wei, Wei-Qi and Bastarache, Lisa A and Carroll, Robert J and Marlo, Joy E and Osterman, Travis J and Gamazon, Eric R and Cox, Nancy J and Roden, Dan M and Denny, Joshua C},
	journal = {Plo{S} One},
	number = {7},
	pages = {e0175508},
	publisher = {Public Library of Science San Francisco, CA USA},
	title = {Evaluating phecodes, clinical classification software, and ICD-9-CM codes for phenome-wide association studies in the electronic health record},
	volume = {12},
	year = {2017}}

@article{yu2015toward,
	author = {Yu, Sheng and Liao, Katherine P and Shaw, Stanley Y and Gainer, Vivian S and Churchill, Susanne E and Szolovits, Peter and Murphy, Shawn N and Kohane, Isaac S and Cai, Tianxi},
	journal = {Journal of the American Medical Informatics Association},
	number = {5},
	pages = {993--1000},
	publisher = {Oxford University Press},
	title = {Toward high-throughput phenotyping: unbiased automated feature extraction and selection from knowledge sources},
	volume = {22},
	year = {2015}}

@article{choi2018mime,
	author = {Choi, Edward and Xiao, Cao and Stewart, Walter and Sun, Jimeng},
	journal = {Advances in Neural Information Processing Systems},
	title = {Mime: Multilevel medical embedding of electronic health records for predictive healthcare},
	volume = {31},
	year = {2018}}

@article{xiong2023knowledge,
	author = {Xiong, Xin and Sweet, Sara Morini and Liu, Molei and Hong, Chuan and Bonzel, Clara-Lea and Panickan, Vidul Ayakulangara and Zhou, Doudou and Wang, Linshanshan and Costa, Lauren and Ho, Yuk-Lam and others},
	journal = {MedRxiv},
	pages = {2023--09},
	publisher = {Cold Spring Harbor Laboratory Press},
	title = {Knowledge-driven online multimodal automated phenotyping system},
	year = {2023}}

@article{bodenreider2004unified,
	author = {Bodenreider, Olivier},
	journal = {Nucleic Acids Research},
	number = {suppl\_1},
	pages = {D267--D270},
	publisher = {Oxford University Press},
	title = {The unified medical language system ({UMLS}): integrating biomedical terminology},
	volume = {32},
	year = {2004}}

@article{hong2021clinical,
	author = {Hong, Chuan and Rush, Everett and Liu, Molei and Zhou, Doudou and Sun, Jiehuan and Sonabend, Aaron and Castro, Victor M and Schubert, Petra and Panickan, Vidul A and Cai, Tianrun and others},
	journal = {NPJ Digital Medicine},
	number = {1},
	pages = {151},
	publisher = {Nature Publishing Group UK London},
	title = {Clinical knowledge extraction via sparse embedding regression (KESER) with multi-center large scale electronic health record data},
	volume = {4},
	year = {2021}}

@article{williams2000using,
	author = {Williams, Christopher and Seeger, Matthias},
	journal = {Advances in Neural Information Processing Systems},
	title = {Using the Nystr{\"o}m method to speed up kernel machines},
	volume = {13},
	year = {2000}}

@inproceedings{scholkopf1997kernel,
	author = {Sch{\"o}lkopf, Bernhard and Smola, Alexander and M{\"u}ller, Klaus-Robert},
	booktitle = {International Conference on Artificial Neural Networks},
	organization = {Springer},
	pages = {583--588},
	title = {Kernel principal component analysis},
	year = {1997}}

@article{chen2012sparse,
	author = {Chen, Lisha and Huang, Jianhua Z},
	journal = {Journal of the American Statistical Association},
	number = {500},
	pages = {1533--1545},
	publisher = {Taylor \& Francis},
	title = {Sparse reduced-rank regression for simultaneous dimension reduction and variable selection},
	volume = {107},
	year = {2012}}

@article{chen2023statistical,
	author = {Chen, Yunxiao and Li, Chengcheng and Ouyang, Jing and Xu, Gongjun},
	journal = {Journal of Machine Learning Research},
	number = {95},
	pages = {1--66},
	title = {Statistical inference for noisy incomplete binary matrix},
	volume = {24},
	year = {2023}}

@article{wang2022maximum,
	author = {Wang, Fa},
	journal = {Journal of Econometrics},
	number = {1},
	pages = {180--200},
	publisher = {Elsevier},
	title = {Maximum likelihood estimation and inference for high dimensional generalized factor models with application to factor-augmented regressions},
	volume = {229},
	year = {2022}}

@article{chen2019joint,
	author = {Chen, Yunxiao and Li, Xiaoou and Zhang, Siliang},
	journal = {Psychometrika},
	number = {1},
	pages = {124--146},
	publisher = {Cambridge University Press \& Assessment},
	title = {Joint maximum likelihood estimation for high-dimensional exploratory item factor analysis},
	volume = {84},
	year = {2019}}

@article{Chatterjee2015-hs,
	author = {Chatterjee, Sourav},
	doi = {10.1214/14-AOS1272},
	issn = {0090-5364},
	journal = {The Annals of Statistics},
	number = 1,
	pages = {177--214},
	publisher = {The Institute of Mathematical Statistics},
	title = {{Matrix estimation by Universal Singular Value Thresholding}},
	volume = 43,
	year = 2015}

@article{wu2024general,
	author = {Wu, Shihao and Xu, Gongjun and Zhu, Ji},
	journal = {arXiv preprint arXiv:2410.12108},
	title = {A General Latent Embedding Approach for Modeling Non-uniform High-dimensional Sparse Hypergraphs with Multiplicity},
	year = {2024}}

@article{park2018finding,
	author = {Park, Dohyung and Kyrillidis, Anastasios and Caramanis, Constantine and Sanghavi, Sujay},
	journal = {SIAM Journal on Imaging Sciences},
	number = {4},
	pages = {2165--2204},
	publisher = {SIAM},
	title = {Finding low-rank solutions via nonconvex matrix factorization, efficiently and provably},
	volume = {11},
	year = {2018}}

@article{Braun2006-rc,
	author = {Braun, M},
	issn = {1533-7928,1532-4435},
	journal = {Journal of Machine Learning Research},
	number = 82,
	pages = {2303--2328},
	title = {{Accurate error bounds for the eigenvalues of the kernel matrix}},
	volume = 7,
	year = 2006}

@article{Koltchinskii2000-pz,
	author = {Koltchinskii, Vladimir and Gin{\'e}, Evarist and Gine, Evarist},
	doi = {10.2307/3318636},
	issn = {1350-7265,1573-9759},
	journal = {Bernoulli},
	number = 1,
	pages = 113,
	publisher = {JSTOR},
	title = {{Random matrix approximation of spectra of integral operators}},
	volume = 6,
	year = 2000}

@article{kimeldorf1970correspondence,
	author = {Kimeldorf, George S and Wahba, Grace},
	journal = {The Annals of Mathematical Statistics},
	number = {2},
	pages = {495--502},
	publisher = {JSTOR},
	title = {A correspondence between Bayesian estimation on stochastic processes and smoothing by splines},
	volume = {41},
	year = {1970}}

@article{mikolov2013exploiting,
	author = {Mikolov, Tomas and Le, Quoc V and Sutskever, Ilya},
	journal = {arXiv preprint arXiv:1309.4168},
	title = {Exploiting similarities among languages for machine translation},
	year = {2013}}

@article{Shi2021-xe,
	author = {Shi, Xu and Li, Xiaoou and Cai, Tianxi},
	doi = {10.1080/01621459.2020.1752219},
	issn = {0162-1459},
	journal = {Journal of the American Statistical Association},
	number = 536,
	pages = {1953--1964},
	publisher = {Taylor \& Francis},
	title = {{Spherical Regression Under Mismatch Corruption With Application to Automated Knowledge Translation}},
	volume = 116,
	year = 2021}

@article{gan2025arch,
	author = {Gan, Ziming and Zhou, Doudou and Rush, Everett and Panickan, Vidul A and Ho, Yuk-Lam and Ostrouchovm, George and Xu, Zhiwei and Shen, Shuting and Xiong, Xin and Greco, Kimberly F and others},
	journal = {Journal of Biomedical Informatics},
	pages = {104761},
	publisher = {Elsevier},
	title = {Arch: Large-scale knowledge graph via aggregated narrative codified health records analysis},
	volume = {162},
	year = {2025}}

@article{chen2018robust,
	author = {Chen, Yunxiao and Li, Xiaoou and Liu, Jingchen and Ying, Zhiliang},
	journal = {Psychometrika},
	pages = {538--562},
	publisher = {Springer},
	title = {Robust measurement via a fused latent and graphical item response theory model},
	volume = {83},
	year = {2018}}

@article{she2019cross,
	author = {She, Yiyuan and Tran, Hoang},
	journal = {Journal of the Royal Statistical Society: Series B (Statistical Methodology)},
	number = {1},
	pages = {145--161},
	publisher = {Wiley Online Library},
	title = {On cross-validation for sparse reduced rank regression},
	volume = {81},
	year = {2019}}

@article{bunea2011optimal,
	author = {Bunea, Florentina and She, Yiyuan and Wegkamp, Marten H},
	journal = {The Annals of Statistics},
	number = {2},
	pages = {1282--1309},
	publisher = {Institute of Mathematical Statistics},
	title = {Optimal selection of reduced rank estimators of high-dimensional matrices},
	volume = {39},
	year = {2011}}

@article{park2024low,
	author = {Park, Seyoung and Lee, Eun Ryung and Zhao, Hongyu},
	journal = {Journal of the American Statistical Association},
	number = {545},
	pages = {202--216},
	publisher = {Taylor \& Francis},
	title = {Low-rank regression models for multiple binary responses and their applications to cancer cell-line encyclopedia data},
	volume = {119},
	year = {2024}}

\end{document}